\documentclass{article}

\usepackage{microtype}
\usepackage{graphicx}
\usepackage{subcaption}
\usepackage{colortbl,booktabs} 
\usepackage{multicol}
\usepackage{multirow}
\usepackage{threeparttable}

\usepackage{hyperref}

\usepackage[accepted]{icml2023}

\usepackage{amsmath}
\usepackage{amssymb}
\usepackage{mathtools}
\usepackage{amsthm}
\usepackage{mathrsfs}

\usepackage[capitalize,noabbrev]{cleveref}

\theoremstyle{plain}
\newtheorem{theorem}{Theorem}[section]

\newtheorem{lemma}[theorem]{Lemma}

\theoremstyle{definition}
\newtheorem{definition}[theorem]{Definition}

\theoremstyle{remark}

\usepackage{amsmath,amsfonts,bm}
\usepackage{amsthm, amssymb, bbm}
\usepackage{thm-restate}
\usepackage{thmtools}
\usepackage{mathtools}

\def\eqref#1{equation~\ref{#1}}

\def\1{\bm{1}}

\DeclareMathAlphabet{\mathsfit}{\encodingdefault}{\sfdefault}{m}{sl}
\SetMathAlphabet{\mathsfit}{bold}{\encodingdefault}{\sfdefault}{bx}{n}

\DeclareMathOperator*{\argmin}{arg\,min}
\DeclareMathOperator*\bigcircop{\bigcirc}

\newcommand{\etal}{{et~al.}\ }

\newcommand{\norm}[2]{||#1||_{#2}}

\usepackage{microtype}
\usepackage{mdframed}
\mdfdefinestyle{MyFrame}{%
    linecolor=gray,
    innertopmargin=2pt,
    innerbottommargin=2pt,
    innerrightmargin=2pt,
    roundcorner=10pt,
    innerleftmargin=2pt,
    linewidth=1pt,
    backgroundcolor=white
    }

\newboolean{showcomments}
\setboolean{showcomments}{true}
\ifthenelse{\boolean{showcomments}}
 { \newcommand{\mynote}[2]{
      \fbox{\bfseries\sffamily\scriptsize#1}
        {\small$\blacktriangleright$\textsf{\emph{#2}}$\blacktriangleleft$}}}
        { \newcommand{\mynote}[2]{}}

\definecolor{DarkOrange}{rgb}{0.8,0.3,0.0} 
\definecolor{DarkCyan}{rgb}{0.0, 0.55, 0.55}
\definecolor{Gray}{gray}{0.9}

\usepackage[textsize=tiny]{todonotes}

\icmltitlerunning{GAT: Guided Adversarial Training with Pareto-optimal Auxiliary Tasks}

\begin{document}

\twocolumn[
\icmltitle{GAT: Guided Adversarial Training with Pareto-optimal Auxiliary Tasks}




\begin{icmlauthorlist}
\icmlauthor{Salah Ghamizi}{uni}
\icmlauthor{Jingfeng Zhang}{riken}
\icmlauthor{Maxime Cordy}{uni}
\icmlauthor{Mike Papadakis}{uni}
\icmlauthor{Masashi Sugiyama}{riken,todai}
\icmlauthor{Yves Le Traon}{uni}
\end{icmlauthorlist}

\icmlaffiliation{uni}{The University of Luxembourg}
\icmlaffiliation{riken}{RIKEN Center for Advanced Intelligence
Project (AIP)}
\icmlaffiliation{todai}{The University of Tokyo}

\icmlcorrespondingauthor{Jingfeng Zhang}{jingfeng.zhang@riken.jp}

\icmlkeywords{multi-task, adversarial, robustness, semi-supervised, multi-objective}

\vskip 0.3in
]



\printAffiliationsAndNotice{}  

\begin{abstract}

While leveraging additional training data is well established to improve adversarial robustness, it incurs the unavoidable cost of data collection and the heavy computation to train models. 
To mitigate the costs, we propose \textit{Guided Adversarial Training } (GAT), a novel adversarial training technique that exploits auxiliary tasks under a limited set of training data. 
Our approach extends single-task models into multi-task models during the min-max optimization of adversarial training, and drives the loss optimization with a regularization of the gradient curvature across multiple tasks.
GAT leverages two types of auxiliary tasks: self-supervised tasks, where the labels are generated automatically, and domain-knowledge tasks, where human experts provide additional labels. 
Experimentally, GAT increases the robust AUC of CheXpert medical imaging dataset from 50\% to 83\% and On CIFAR-10, GAT outperforms eight state-of-the-art adversarial training and achieves 56.21\% robust accuracy with Resnet-50. 
Overall, we demonstrate that guided multi-task learning is an actionable and promising avenue to push further the boundaries of model robustness.

\end{abstract}

\section{Introduction}

Despite their impressive performance, Deep Neural Networks (DNNs) are sensitive to small, imperceptible perturbations in the input. The resulting \emph{adversarial inputs} raise multiple questions about the robustness of such systems, especially in safety-critical domains such as autonomous driving \cite{cao2019adversarial}, financial services \cite{coeva2}, and medical imaging \cite{understandingadvattacks}.


Adversarial training (AT) \cite{madry2017towards} is the de facto standard for building robust models. In its simplest form, AT trains the model with the original training data and adversarial examples crafted from them. Robustness can be further increased with additional data in the AT process, including unlabeled data \cite{carmon2019unlabeled}, augmented data \cite{rebuffi2021data}, or artificial data from generative models \cite{gowal2021improving}. These approaches produce significantly more robust models (e.g. ResNet50 with 51.56\% robust accuracy on CIFAR-10) and can be further enhanced through the use of very large models (WideResNet-70-16 with 66\% robust accuracy) \cite{croce2020robustbench}.


However, the robustness achieved by AT with data augmentation has already reached a plateau \cite{schmidt2018adversarially, gowal2021improving} whereas the computational costs of AT in large models prohibit their use at scale. This is why research has explored new techniques to increase robustness, taking inspiration from, e.g., Neural Architecture Search ~\cite{nas_splc,nas_cvpr} and self-supervised learning~\cite{hendrycks2019using,chen2020adversarial}. They are, however, not yet competitive to data augmentation techniques in terms of clean and robust performances.


In this paper, we propose \textbf{Guided Adversarial Training (GAT)}, a new technique based on multi-task learning to improve AT. Inspired from preliminary investigations of robustness in multi-task models \citet{mao2020multitask} and \citet{ghamizi2022mt}, we demonstrate that robustness can be improved by adding auxiliary tasks to the model and introducing a gradient curvature minimization and a multi-objective weighting strategy into the AT optimization process. Our novel regularization can  achieve optimal pareto-fronts across the tasks for both clean and robust performances.
To this end, GAT can exploit both self-supervised tasks without human intervention (e.g. image rotation) and domain-knowledge tasks using human-provided labels.

Our experiments demonstrate that GAT outperforms eight state-of-the-art AT techniques based on data augmentation and training optimization, with an improvement on CIFAR-10 of 3.14\% to 26.4\% compared to state of the art adversarial training with data-augmentation. GAT shines in scarce data scenarios (e.g. medical diagnosis tasks), where data augmentation is not applicable.

Our large study across five datasets and six tasks demonstrates that task augmentation is an efficient alternative to data augmentation, and can be key to achieving both clean and robust performances.

Our algorithm and replication packages are available on \href{https://github.com/yamizi/taskaugment}{https://github.com/yamizi/taskaugment}

\section{Background}
\label{sec:background}

\subsection{Multi-task learning (MTL)}

MTL leverages shared knowledge across multiple tasks to learn models with higher efficiency \cite{vandenhende2021multi,standley2020tasks}. 
A multi-task model is commonly composed of an encoder that learns shared parameters across the tasks and a decoder part that branches out into task-specific heads. 

We can view MTL as a form of inductive bias. By introducing an inductive bias, MTL causes a model to prefer some hypotheses over others \cite{ruder2017overview}. 
MTL effectively increases the sample size we are using to train our model. As different tasks have different noise patterns, a model that learns two tasks simultaneously can learn a more general representation. Learning task A alone bears the risk of overfitting to task A, while learning A and B jointly enables the model to obtain a better representation \cite{caruana1997multitask}.

\subsection{Adversarial robustness}

An adversarial attack is the process of intentionally introducing perturbations on the inputs of a model to cause the wrong predictions. 
One of the earliest attacks is the Fast Gradient Sign Method (FGSM) \cite{fgsm_goodfellow2015explaining}. It adds a small perturbation \(\delta\) to the input of a neural network, which is defined as: $\eta = \epsilon \, \text{sign}(\delta_x \mathscr{L}(\theta,x,y))$,
where \(\theta\) are the parameters of the network, \(x\) is the input data, \(y\) is its associated label, \(\mathscr{L}_\theta(x,y)\) is the loss function used, and \(\epsilon\) is the strength of the attack.

Following \citet{fgsm_goodfellow2015explaining}, other attacks were proposed, such as by adding iterations (I-FGSM)~\cite{kurakin2016adversarial}, projections and random restart (PGD)~\cite{pgd_madry2019deep}, and momentum (MIM)~\cite{dong2018boosting}.

Given a multi-task model $\mathscr{M}_\theta$ parameterized by $\theta$ for $M$ tasks, an input example $x$, and its corresponding ground-truth label $\bar{y}$, 
the attacker seeks the perturbation $\delta$ that will maximize the joint loss $\mathscr{L}_\theta$ of the attacked tasks:

\begin{equation}
 \underset{\delta \in \Delta}{\operatorname{argmax}} \: \mathscr{L}_\theta(x+\delta,\bar{y}) \: \text{s.t.} \: \norm{\delta}{p} \leq \epsilon ,
\label{eq:multi-task-objective}
\end{equation}

where $p \in \{1,2,\infty\}$ and $\|\cdot\|_p$ denotes the $\ell_p$-norm. 
A typical choice for a perturbation space is to take $\Delta = \{\delta : \|\delta\|_\infty \leq \epsilon\}$ for some $\epsilon >0$. 

$\mathscr{L}_\theta(x+\delta,\bar{y}) = \sum_{j=1}^{M} \mathscr{L}_j(x+\delta, y_{j})$ is the joint loss of the M attacked tasks.

\paragraph{Adversarial training (AT)}

AT is a method for learning networks which are robust to adversarial attacks. Given a multi-task model $\mathscr{M}_\theta$ parameterized by $\theta$, a dataset $\{ (x_i, y_i) \}$, a loss function $\mathscr{L}_\theta$, and a perturbation space $\Delta$
, the learning problem is cast as the following optimization:

\begin{equation}
    \min_\theta \sum_i \max_{\delta \in \Delta} \mathscr{L}_\theta(x_i + \delta, y_i)
    \label{eq:adv_training}
\end{equation}

The procedure for AT uses some adversarial attack to approximate the inner maximization over $\Delta$, followed by some variation of gradient descent on the model parameters $\theta$. 

\subsection{Adversarial vulnerability}

\citet{simon2019first} introduced the concept of \textbf{adversarial vulnerability} to evaluate and compare the robustness of single-task models and settings. \citet{mao2020multitask} extended it to multi-task models as follows:

\begin{definition}

Let $\mathscr{M}$ be a multi-task model, $\mathscr{T}' \subseteq \mathscr{T}$ be a subset of its tasks, and $\mathcal{L}_{\mathscr{T}'}$ be the joint loss of tasks in $\mathscr{T}'$. Then, we denote by $\mathbb{E}_{x}[\delta\mathcal{L}(\mathscr{T}',\epsilon)]$ the \emph{adversarial vulnerability} of $\mathscr{M}$ on $\mathscr{T}'$ to an $\epsilon$-sized $\| .\|_p$-attack, and define it as the average increase of $\mathcal{L}_{\mathscr{T}'}$ after attack over the whole dataset:

$$    \mathbb{E}_{x}[\delta \mathcal{L}(\mathscr{T}',\epsilon)]= \\ \mathbb{E}_{x}\left[\max _{\|\delta\|_{p}\leq\epsilon} \mid \mathcal{L}_{\mathscr{T}'}(x+\delta, \bar{y})-\mathcal{L}_{\mathscr{T}'}(x, \bar{y}) \mid \right]
$$
\end{definition}

This definition matches the definitions of previous work \cite{goodfellow2015explaining,sinha2017certifying} of the robustness of deep learning models: the models are considered vulnerable when a small perturbation causes a large average variation of the joint loss.

Similarly, \emph{the adversarial task vulnerability} of a task $i$ is the average increase of $\mathcal{L}_{\mathscr{T}'}(x,y_i)$ after attack.

\section{Preliminaries}
\label{sec:preliminaries}


To build our method, we first investigate the factors that influence the robustness of multi-task models. This preliminary study enables us to derive relevant metrics to optimize during the AT process for multi-task models.

Our idea stems from previous observations that task weights can have a significant impact on the 
robustness of multi-task models \cite{Ghamizi_Cordy_Papadakis_Traon_2022}. We pursue this investigation and identify the three main factors of adversarial vulnerability in multi-task models: The relative orientation of the gradient of the tasks' loss, their magnitude similarity and the weighting of the clean and adversarial contributions to the loss (Proof in Appendix \ref{appendixA1}).

\begin{theorem}

Consider a multi-task model $\mathscr{M}$ where an attacker targets $\mathscr{T}= \{t_1,t_2\}$ two tasks weighted with $\alpha_1$ and $\alpha_2$ respectively, with an $\epsilon$-sized $\| .\|_p$-attack. If the model is converged, and the gradient for each task is i.i.d. with zero mean and the tasks are correlated, the adversarial vulnerability of the model can be approximated as

\begin{equation}
\begin{aligned}
    \mathbb{E}_{x}[\delta \mathcal{L}'] \propto \sqrt{1+ 2 \frac{\alpha_1 . \alpha_2 . \operatorname{Cov}\left(\partial_{x} \mathcal{L}_1, \partial_{x} \mathcal{L}_2\right)}{\alpha_1^2 \sigma_1^2+ \alpha_2^2 \sigma_2^2 }},
\end{aligned}
\end{equation}
\label{theorem3}
\end{theorem} 

where $ \sigma_i^2 = \operatorname{Cov}\left(\partial_{x} \mathcal{L}_i, \partial_{x} \mathcal{L}_i\right)$ and $\partial_{x} \mathcal{L}(x, y_i)$ the gradient of the task $i$.

The above theorem reveals that adversarial vulnerability is particularly sensitive to the relative amplitude of the gradients of the tasks and their orientation.
In standard multi-task learning (MTL), the relative properties of the task gradients -- such as the orientation angle, magnitude similarity, and curvature -- have an impact on the learning speed and on the achieved clean accuracy \cite{vandenhende2021multi}. Therefore, task weighting approaches like  \emph{Projecting Conflicting Gradients (PCG)} \cite{weight_pcg} rely on these properties to optimize standard training.


\subsection{Empirical study}
\label{sec:emipirical_study}
To confirm empirically the findings of Theorem \ref{theorem3}, we study the following metrics and empirically check their correlation to robustness.

\begin{definition}
\label{def:angle}
Let $\phi_{ij}$ be the angle between two tasks' gradients $\mathbf{g}_i$ and $\mathbf{g}_j$. We define the gradients as \textbf{conflicting} when $\cos\phi_{ij}<0$.
\end{definition}

\begin{definition}
\label{def:div}
The \textbf{gradient magnitude similarity} between two gradients $\mathbf{g}_i$ and $\mathbf{g}_j$
is 
$
    \Phi(\mathbf{g}_i, \mathbf{g}_j) = \frac{2\|\mathbf{g}_i\|_2\|\mathbf{g}_j\|_2}{\|\mathbf{g}_i\|_2^2 + \|\mathbf{g}_j\|_2^2}.
$
\end{definition}
When the magnitude of two gradients is the same, this value equals 1. As the gradient magnitude difference increases, the similarity goes towards zero.

\begin{definition}
\label{def:curvature}
The \textbf{multi-task curvature bounding measure} between two gradients $\mathbf{g}_i$ and $\mathbf{g}_j$ is
$\xi(\mathbf{g}_i, \mathbf{g}_j) = (1 - \cos^2\phi_{ij})\frac{\|\mathbf{g}_i - \mathbf{g}_j|_2^2}{\|\mathbf{g}_i + \mathbf{g}_j\|_2^2}.$
\end{definition}

The multi-task curvature bounding measure combines information about both the orientation of the gradients of the tasks and the relative amplitude of the gradients.

\begin{figure*}[t]
     \centering
     \begin{subfigure}[b]{0.30\textwidth}
         \centering
         \includegraphics[width=\linewidth]{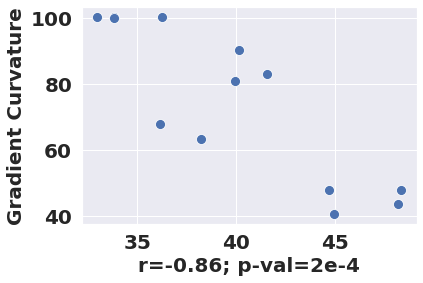}
         
     \end{subfigure}
     \hfill
     \begin{subfigure}[b]{0.30\textwidth}
         \centering
         \includegraphics[width=\linewidth]{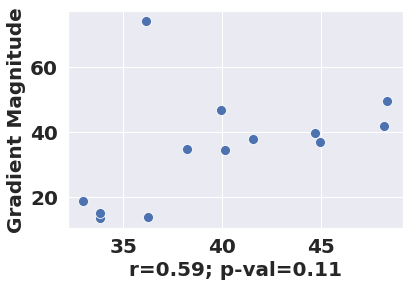}
         
     \end{subfigure}
     \hfill
     \begin{subfigure}[b]{0.30\textwidth}
         \centering
         \includegraphics[width=\linewidth]{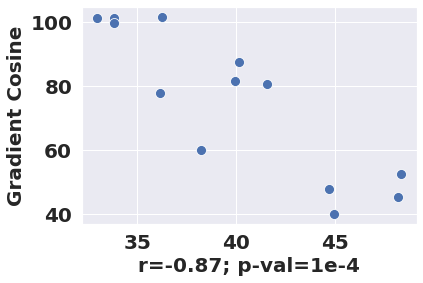}
         
     \end{subfigure}
     \hfill
     \begin{subfigure}[b]{0.30\textwidth}
         \centering
         \includegraphics[width=\linewidth]{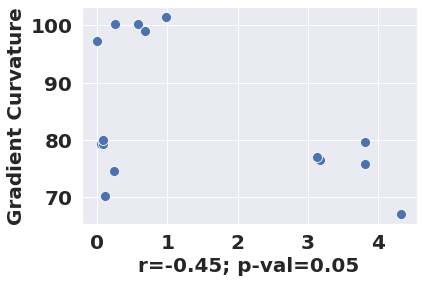}
         
     \end{subfigure}
     \hfill
     \begin{subfigure}[b]{0.30\textwidth}
         \centering
         \includegraphics[width=\linewidth]{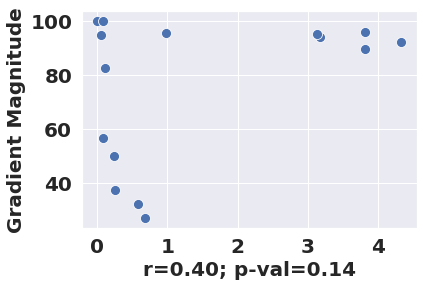}
         
     \end{subfigure}
     \hfill
     \begin{subfigure}[b]{0.30\textwidth}
         \centering
         \includegraphics[width=\linewidth]{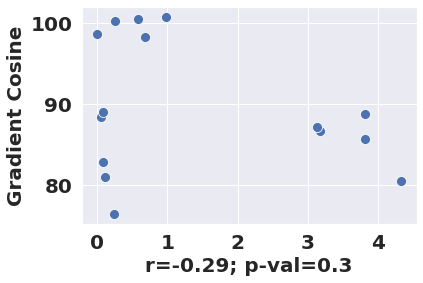}

     \end{subfigure}
     
     \hfill
     \vspace{-5mm}
             \caption{Robust accuracy (X-axis) with each of our three metrics (Y-axis). Top: Models with AT, bottom: Models with standard training. Left: Gradient multi-task curvature bounding measure, middle: Gradient magnitude similarity, right: Gradient cosine angle. Below each scatter plot is the Pearson correlation coefficient r and its p-value between the robust accuracy and the studied metric. This study confirms that the Gradient Curvature is a good surrogate for adversarial robustness and can be used to optimize the robustness.}
        \label{fig:correlation_metrics}
        
\end{figure*}

We evaluate in Fig. \ref{fig:correlation_metrics} the Pearson correlation coefficient between the robust accuracy and each of the three metrics. 
For adversarially trained models (top), both the \textbf{Gradient multi-task curvature bounding measure} (left) and the \textbf{Gradient cosine angle} (right) are strongly negatively correlated with the adversarial robustness, with respectively a correlation coefficient $r$ of $-0.86$ and $-0.87$. However, for models trained with standard training, only the \textbf{Gradient multi-task curvature bounding measure} is negatively correlated ($r=-0.45$) to the robustness of the models.

These results show that the \textbf{gradient curvature measure} can be a good surrogate to study the robustness of MTL models, especially with AT. The negative correlation between the gradient curvature measure and the robust accuracy suggests that a flatter multitask loss landscape leads to more robust models. Our results are in coherence with the seminal works from \citet{engstrom} and \citet{moosavi} studied in the single-task setting.

\section{Method}

Based on our preliminary findings, we propose GAT -- Guided Adversarial Training -- as a new approach for effective AT. GAT introduces three novel components: (1) a multi-task AT using both self-supervised and domain-knowledge tasks, (2) a gradient curvature regularization that guides the AT towards less vulnerable loss landscapes, and (3) a pareto-optimal multi-objective optimization of the weights of each loss (clean and robust losses for target tasks) at each step of the min-max optimization of AT. 

\subsection{The proposed approach: GAT}
\label{sec:approach}

\begin{figure*}

\begin{center}
\centerline{\includegraphics[width=2\columnwidth]{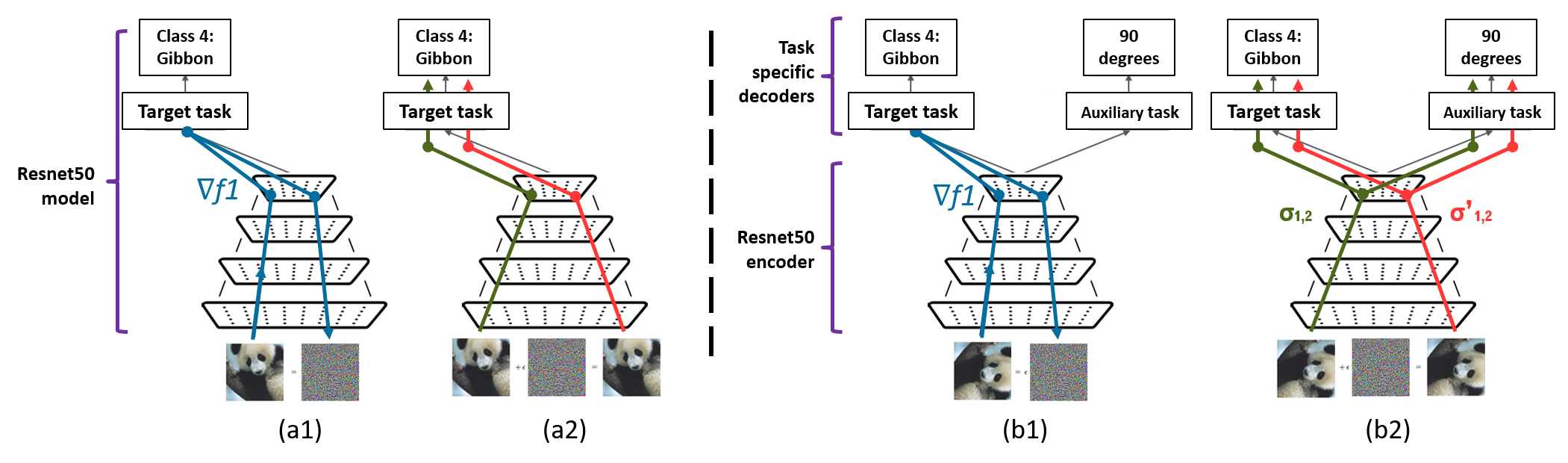}}
\vspace{-5mm}
\caption{Comparison of single-task AT (a) and our proposed approach GAT (b). GAT preserves the original target task and adds an auxiliary task where abundant labels are available: For instance, a self-supervised task like rotation angle prediction.  
In (a1) and (b1), we generate the adversarial example using only the loss of the target task (blue line). We update the models' weights with backpropagation in (a2) and (b2). We compute the model's weights update with GAT (b2) using a weighted combination ($\sigma_{1,2}, \sigma'_{1,2}$) of the loss of the different tasks over the clean examples (green line) and the adversarial examples (red line).
}

\label{fig:overview}
\end{center}
\end{figure*}

GAT first transforms any single-task model into a multi-task model before AT. We connect additional decoders to the penultimate layer of the existing model. The architecture of each decoder is selected for one auxiliary task specifically. For example, we use a single dense layer as a decoder for classification tasks and a U-net \cite{unet} decoder for segmentation tasks.
In Figure \ref{fig:overview}, we extend an ImageNet classification model into a multi-task model that learns both the class (target task) and the orientation (auxiliary task) of the image. The auxiliary task here is 'the rotation angle prediction', a self-supervised classification task where we can generate the labels on the fly by rotating the original image. 

We consider two types of task augmentation. In \emph{self-supervised} task augmentation, the image is pre-processed with some image transformation like jigsaw scrambling \cite{noroozi2016jigsaw} or image rotation \cite{gidaris2018rotation}. The auxiliary task predicts the applied image transformation (e.g., the permutation matrix for the jigsaw task, the rotation angle for the rotation task). In \emph{domain-knowledge} tasks, a human oracle provides additional labels. In the medical imaging case, these additional labels may include, e.g., other pathologies and stages or patient data like gender and age.

A naive implementation of AT (Eq.~(\ref{eq:adv_training})) to multi-task models consists of the following min-max optimization problem: 
\begin{equation}
    \min_\theta \sum_i \max_{\delta \in \Delta} \sum_{j=1}^{M} (\mathscr{L}_j(x_i, y_{i,j}) + \mathscr{L}_j(x_i + \delta, y_{i,j}) ),
    \label{eq:GAT}
\end{equation}
where 
$y_{i,j}$ is the label of the input example $i$ for the task $j$.

\subsection{Guiding the gradient curvature with regularization}

We showed in Section \ref{sec:preliminaries} that the curvature bounding measure is a reliable surrogate of the adversarial robustness of models. Hence, we guide the AT of Eq.~(\ref{eq:GAT}) with a curvature regularization term: $$\mathscr{L}_{j}^\mathrm{(reg)} = \sum_{k=1}^{j-1} (1 - \cos^2\phi_{jk})\frac{\|\mathbf{g}_j - \mathbf{g}_k|_2^2}{\|\mathbf{g}_j + \mathbf{g}_k\|_2^2}.$$

Therefore, the definitive formulation of GAT:
\begin{equation}
\label{eq:WGAT}
\begin{array}{l}
    \min_\theta \sum_i \max_{\delta \in \Delta} \sum_{j=1}^{M} \\
    \left(\alpha_{j}^\mathrm{(clean)} \mathscr{L}_j(x_i, y_{i,j}) + \alpha_{j}^\mathrm{(adv)} \mathscr{L}_j(x_i + \delta, y_{i,j})  + \mathscr{L}_{j}^\mathrm{(reg)} \right)

\end{array}
\end{equation}
where $\alpha_{j}^\mathrm{(clean)}$ and $\alpha_{j}^\mathrm{(adv)}$ are positive weights that control the relative contribution of the clean and adversarial loss (respectively) of task $j$ to the objective function to optimize. Their value will be optimized automatically throughout the AT process, as proposed below.

\subsection{Adversarial Training as a multi-objective optimization problem (MOOP)}

The optimization proposed in Eq.~(\ref{eq:WGAT}) faces conflicting gradients between the clean and adversarial losses, and possibly between the target and auxiliary tasks.
Weighting strategies for MTL \cite{weight_imtl, weight_pcg, weight_gv} all assume that the tasks' gradients are misaligned and not totally opposed. The case of GAT is more complex because there is no guarantee that this assumption holds across the AT optimization. Instead of achieving the minimization of the whole loss, we seek to reach a Pareto-stationary point where we cannot improve the loss of one task without degrading the loss of another task \cite{Kais99}. 

To solve this MOOP, we extend the Multi-Gradient Descent Algorithm (MGDA)~\cite{desideri_mgda} to AT. We generalize gradient descent to multi-objective settings by identifying a descent direction common to all objectives (i.e., clean and robust losses of target tasks) and tune the weights of the tasks' losses at each adversarial training batch. MGDA formally guarantees convergence to a Pareto-stationary point \cite{desideri_mgda} to achieve both clean and robust performances.
Subsequent research by \citet{weight_mgda} has shown that the Multi-Gradient Descent Algorithm (MGDA) "yields a Pareto optimal solution under realistic assumptions". We rely on these assumptions and the upper bound they propose using the Frank-Wolfe-based optimizer (Algorithm \ref{alg:mtl_mgda_solver}) to achieve Pareto optimality. The only assumption we use is the same as the one proven by \citet{weight_mgda}, namely, the non-singularity assumption.
The assumption is reasonable because the singularity implies that tasks are linearly related, and a trade-off is not necessary. Our empirical study in section \ref{sec:emipirical_study} confirms that our tasks are not linearly related.


Algorithm \ref{algorithm:pseudoalgo} presents GAT and is explained in details (including the MGDA procedure) in Appendix \ref{appendixA2}.

 \begin{algorithm}[t]
\caption{Pseudo-Algorithm of GAT}
\label{algorithm:pseudoalgo}
\small
\begin{algorithmic}

\STATE{\textbf{Given}}: a single task model $\mathscr{M}$ parameterized by $\theta^{s}$ for the shared encoder and $\theta^{t}$ for the specific heads, a batch example $x$, and $\bar{y}= (y_1,..., y_s, ... y_m)$ its corresponding labels for each task, with $y_1$ the target task, $y_{1 < i \leq s}$ the auxiliary self-supervised tasks and $y_{s < i \leq m}$ the auxiliary domain-knowledge tasks; 
\STATE{\textbf{Given}}: an input processing $f_t$ for each auxiliary self-supervised $t$ task with label $y_{1 < t \leq s}$.
\STATE{\textbf{Given}}: a $PGD$ adversarial attack with a step size $\epsilon_{step}$; a maximum perturbation $\epsilon$; $S$ number of attack iterations;

 \STATE{\textbf{Step 1}:} Create a decoder $D_i$ at the penultimate layer of $\mathscr{M}$ for each of the auxiliary task $t_i$ / $i>1$. 
       
\STATE{\textbf{Step 2}:} For each epoch and batch $x$ 
Do
       \begin{enumerate}
           \item For each self-supervised task $t_{1 < i \leq s}$:\newline  $x \leftarrow  \bigcircop_{t=2}^s \ f_{t}(x)$
           \item $\hat{x} \leftarrow \mathrm{PGD}(x, y_1, \epsilon_{step}, \epsilon, S)$.
           \item Get the task losses $l$ and regularization losses $l^{(reg)}$:  \newline $l \leftarrow l_{1,x},l_{1,\hat{x}},\ldots , l_{M,x},l_{M,\hat{x}}, l^{(reg)}_{M+1,x} ,\ldots , l^{(reg)}_{2M-1,x}$.
           \item 
           $\alpha_1,\ldots,\alpha_{2M-1} \leftarrow$ MGDA($\theta$,$l$)
           \item 
           $\theta^{s} \leftarrow \theta^{s} -   \eta \sum_{t=1}^{2M-1} \alpha_{t} \nabla_{\theta^{sh}}  l_{t,x}(\theta^{s},\theta^t)$
       \end{enumerate}
\STATE{\textbf{Step 3}:} Disable the auxiliary branches added at step 1.
\end{algorithmic}

\end{algorithm}
\section{Experiments}
We extensively evaluate GAT on two datasets and demonstrate that GAT achieves better robust performances than SoTA data augmentation AT: We compare GAT with Cutmix \cite{yun2019cutmix}, MaxUp \cite{maxup}, unlabeled data augmentation \cite{carmon2019unlabeled}, 
denoising diffusion probabilistic models augmentations (DDPM) \cite{gowal2021improving}, self-supervised pre-training \cite{chen2020adversarial}, self-supervised multi-task learning \cite{hendrycks2019using}. We also compare GAT to three popular AT approaches that do not focus on data augmentation: Madry adversarial training \cite{pgd_madry2019deep}, TRADES adversarial training \cite{zhang2019trades}, and FAST adversarial training \cite{wong2020fast}. 

The extension of our evaluation to different settings is discussed in Section~\ref{sec:generalization}.

\subsection{Experimental setup}
We present below our main settings. Further details 
are in Appendix \ref{appendixSetup}.

\paragraph{Datasets.}

CIFAR-10 \cite{cifar10} is a 32x32 color image dataset. We evaluate two scenarios: A full 50.000 image AT scenario and an AT scenario using a subset of 10\% to simulate scarce data scenarios. A study with 25\%, and 50\% of the original data is in the Appendix \ref{appendixB1}.  

CheXpert~\cite{irvin2019chexpert} is a public chest X-ray dataset. It consists of 512x512 grayscale image radiography collected from one hospital.
We report below the results for predicting the Edema and the Atelectasis disease as target tasks, and provide results for other combinations of pathologies in  Appendix \ref{appendixB2}. We confirm our results for another medical imaging dataset (NIH) in Appendix \ref{appendixC2}.

\paragraph{Architecture.}

We use an encoder-decoder image classification architecture, with ResNet-50v2 
as encoders for the main study. 
We provide in Appendix \ref{appendixComplementary} complementary studies with WideResnet28 encoders.

\paragraph{Task augmentations.}

For both CIFAR-10 and CheXpert datasets, we evaluate two self-supervised tasks: \textbf{Jigsaw}, where we split the images into 16 chunks and scramble them according to a permutation matrix. The permutation matrix represents the labels of the Jigsaw prediction task. In the \textbf{Rotation} auxiliary task, we rotate the images by 0, 90, 180, or degrees, and the 4 rotation angles are the labels learned by the Rotation prediction task.    

To evaluate domain-knowledge tasks, we generate new labels as follows. For CIFAR-10, we split the existing 10 classes into 2 macro classes: \emph{Vehicles} or \emph{Animals}. We refer to this task as \textbf{Macro}. For CheXpert, we add the binary classification of \textbf{Cardiomegaly} and \textbf{Pneumothorax} as auxiliary tasks. These auxiliary pathologies often co-occur with Edema and Atelectasis. We also extract the age and gender meta-data related to the patients and use them to build auxiliary tasks. Learning the \textbf{Age} is a regression task, while learning the \textbf{Gender} is a 3-class classification task.

\paragraph{Training.}

Both natural and AT is combined with common data augmentations (rotation, cropping, scaling), using SGD with lr=0.1, a cosine annealing, and early stopping. We train CIFAR-10 models for 400 epochs and CheXpert models for 200 epochs.
We perform AT following Madry's approach \cite{madry2017towards} with a 10-steps PGD attack and $\epsilon = 8/255$  size budgets, and we only target the main task to craft the adversarial examples.

\subsection{Results}
\label{sec:results}

\paragraph{GAT improves up to 21\% the robustness of CIFAR-10 models over AT strategies.}

\begin{figure}[t]
\begin{center}
\centerline{\includegraphics[width=\columnwidth]{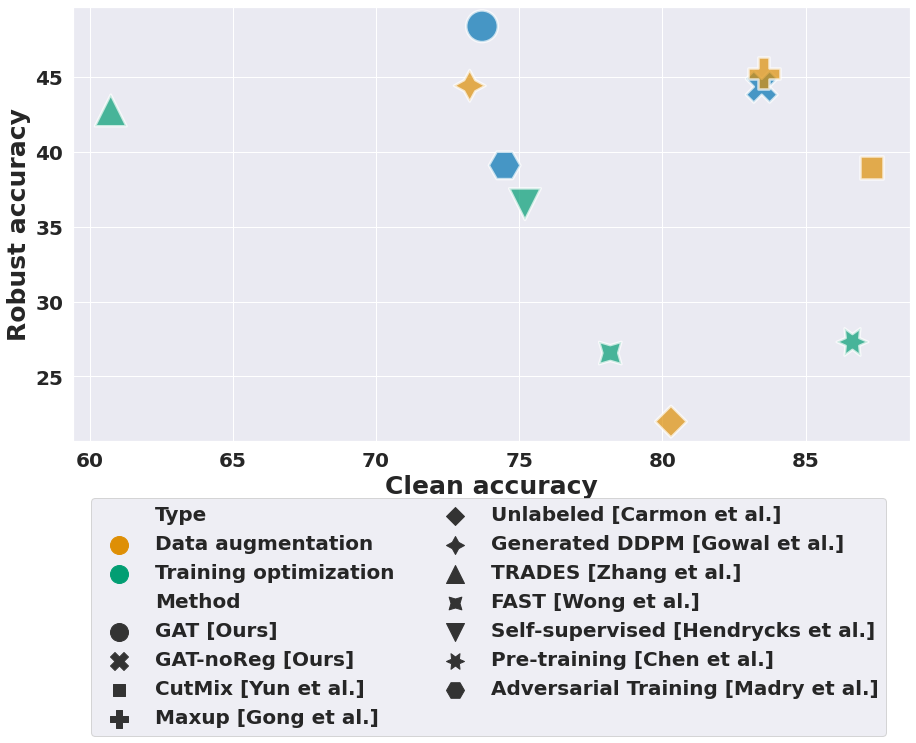}}
\caption{Comparison of GAT performances with state-of-the-art AT mechanisms on CIFAR-10 classification. In yellow AT with data augmentations and in green techniques with AT optimizations.
GAT outperforms all existing approaches in terms of robust accuracy and remains competitive in terms of clean accuracy.}
\label{fig:all_data_cifar}
\end{center}
\end{figure}

We show in Fig. \ref{fig:all_data_cifar} (and Appendix \ref{appendixB3}) the clean and robust accuracy on CIFAR-10 of AT with various optimizations compared to AT with our approach. GAT outperforms data-augmentation AT techniques in terms of robust accuracy from 3.14\% to 26.4\% points, and outperforms AT training optimizations up to 21.82\%.

\paragraph{GAT increases up to 41\% the robustness of medical diagnosis.}

\begin{figure}[t]
\begin{center}
\centerline{\includegraphics[width=\columnwidth]{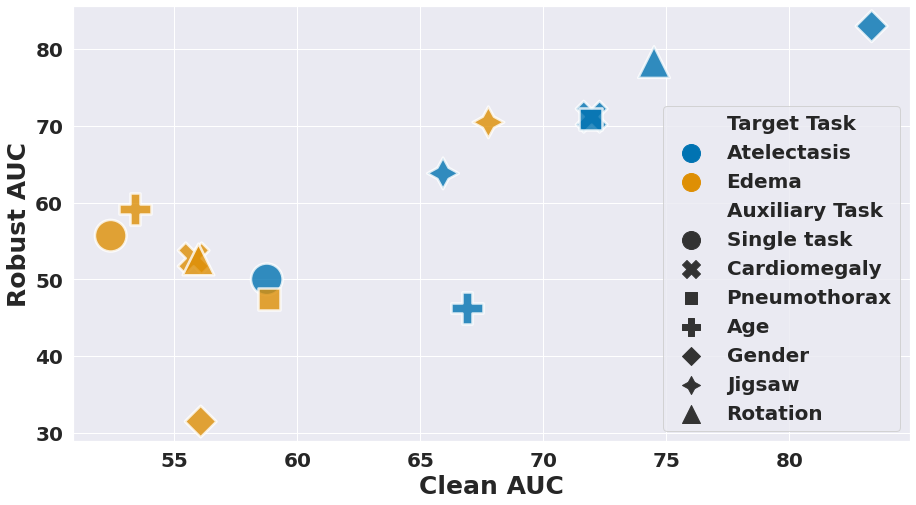}}
\caption{Comparison of different Task Augmentation strategies with single-task models using AT; Clean and robust AUC of GAT vs Single task AT of models trained to diagnose Atelectasis and Edema pathologies.}
\label{fig:limited_data_chex}
\end{center}
\end{figure}

Figure \ref{fig:limited_data_chex} shows the clean and robust AUC of the single-task baseline models (circle marker), and the task-augmented models.

For Atelectasis (blue), \textbf{age} task augmentation leads to lower results than the baseline. However, all remaining task augmentations outperform the baseline both on the clean and robust AUC. The gender augmentation increases the robust AUC of Atelectasis from 58.75\% to 83.34\%. 

For Edema (orange), Task augmentation with Jigsaw leads to the best clean and robust AUC increase. The robust AUC jumps from 55.68\% to 70.47\% compared to single-task AT.

\label{sec:results_aug}



\begin{table*}[]

\caption{The mean and std across three runs for combinations of our approach with data augmentation techniques.  The blue cells indicate the combinations where GAT outperform data augmentation techniques with statistical significance, the underlined cells are the combinations that outperform task augmentation alone and, in bold the best performances.}
\label{tab:sota_cifar10}
\small
\begin{tabular}{l||rrrr|llll}
\hline
Task & \multicolumn{4}{c|}{Robust accuracy (\%)}  & \multicolumn{4}{c}{Clean accuracy (\%)}  \\ 
Augment & None  & Jigsaw& Macro & Rotation  & \multicolumn{1}{r}{None} & \multicolumn{1}{r}{Jigsaw} & \multicolumn{1}{r}{Macro} & \multicolumn{1}{r}{Rotation} \\ \hline
None & 39.09{\small$\pm$0.13} & 32.95{\small$\pm$0.59}& \textbf{48.38}{\small$\pm$0.11} & 36.13{\small$\pm$0.29} & 74.49{\small$\pm$0.16}  & 43.99{\small$\pm$0.36} & 73.70{\small$\pm$0.51} & 56.51{\small$\pm$0.12}\\ \hline
Cutmix   & \cellcolor[HTML]{C0C0C0}38.95{\small$\pm$0.32} & 23.86{\small$\pm$0.55} & \cellcolor[HTML]{34CDF9}41.09{\small$\pm$0.22} & 20.19{\small$\pm$0.04} & \cellcolor[HTML]{C0C0C0}87.31{\small$\pm$0.12}& \underline{60.53}{\small$\pm$1.10} & \underline{87.52}{\small$\pm$0.16} & \cellcolor[HTML]{34CDF9}\underline{\textbf{87.86}}{\small$\pm$0.33}\\ \hline
Unlabeled   & \cellcolor[HTML]{C0C0C0}21.98{\small$\pm$0.18} & \cellcolor[HTML]{34CDF9}26.33{\small$\pm$0.28} & \cellcolor[HTML]{34CDF9}33.88{\small$\pm$0.15} & \cellcolor[HTML]{34CDF9} 35.29{\small$\pm$0.20} & \cellcolor[HTML]{C0C0C0}80.30{\small$\pm$0.41}& \underline{49.32}{\small$\pm$1.06}  & \cellcolor[HTML]{34CDF9}\underline{84.57}{\small$\pm$0.05} & \underline{71.08}{\small$\pm$0.08}\\ \hline
Pre-train& \cellcolor[HTML]{C0C0C0}27.30{\small$\pm$0.40} & \cellcolor[HTML]{34CDF9}\underline{35.47}{\small$\pm$0.38} & \cellcolor[HTML]{34CDF9}32.64{\small$\pm$0.27} & \cellcolor[HTML]{34CDF9}33.78{\small$\pm$0.35} & \cellcolor[HTML]{C0C0C0}86.64{\small$\pm$0.26}& \underline{87.56}{\small$\pm$0.18}  & \underline{86.69}{\small$\pm$0.31} &\underline{87.85}{\small$\pm$0.21}\\ 

\hline
\bottomrule
\end{tabular}
\end{table*}

\paragraph{GAT and data augmentation strategies can be combined to improve clean and robust performances.}

The regularization term of the curvature measure used in GAT involves and may negatively impacts clean performance. We investigate if combining GAT with data-augmentation techniques can mitigate these effects.
We show in Table \ref{tab:sota_cifar10} that GAT with various data augmentation strategies achieves higher robust accuracy than models with data augmentation alone in seven of the nine cases (in blue, Table \ref{tab:sota_cifar10}). The two exceptions are CutMix when combined with \textbf{Jigsaw} or \textbf{Rotation}. Compared to GAT alone, all combinations of GAT with data augmentation show a slight drop in robustness (e.g., GAT with \textbf{Rotation} drops from 36.13\% to 33.78\%) but significantly improve clean accuracy. For example, they increase from 56.51\% to 87.85\% using the \textbf{Rotation} auxiliary task.

\section{Ablation studies}
We demonstrate in the following why the implementation choices of our approach GAT are the best, and how other choices impact the performance of GAT. 

\paragraph{Impact of weighting strategies.}
We proposed to formulate the AT process through the lens of Pareto-stationary optimization. We argue that this Pareto approach is more relevant than other multi-task strategies to handle adversarial and clean losses across multiple (potential) opposing tasks. To confirm this hypothesis, we provide in Table \ref{table:full_adv_pareto} an ablation study of GAT on the CIFAR-10 dataset. Both best clean and robust performances are achieved by GAT. 

Moreover, GAT achieved best the pareto-optimum. We construct the front obtained by each weighting strategy, then compute its associated \emph{hyper-volume} metric. 
This metric measures the volume of the dominated portion of the objective space and lower values indicate better pareto fronts. 
The best values are achieved by GAT with MGDA in Appendix \ref{appendixWeight}.

\begin{table}[t]
\begin{center}
\caption{Ablation study of the impact of the weighting strategies on the test and robust accuracies.}  
\label{table:full_adv_pareto}
\small
\begin{threeparttable}
\begin{tabular}{l||rr}
\hline
                           & Test acc & Robust acc \\
\hline
Macro Task + Equal weights      & 83.00 \%        & 32.16   \%      \\
\hline
Macro Task + MGDA (Ours)   & 73.70 \%        & 48.38   \%      \\
\hline
Macro Task + GV \cite{weight_gv}   & 64.11 \%       & 35.56    \%     \\
\hline
Macro Task + PCG  \cite{weight_pcg} & 63.76 \%        & 44.74   \%     \\
\hline
\bottomrule
\end{tabular}

\end{threeparttable}
\end{center}
\end{table}

\paragraph{Impact of task-dependent adversarial perturbations.}

As our primary focus is the robustness of the main task only, we generate the perturbation $\delta$ only on the main task as presented in step 1 of Figure \ref{fig:overview}. Therefore, the perturbation $\delta$ is independent of the auxiliary tasks in Equation \ref{eq:WGAT}. One can also generate adversarial examples dependent on the auxiliary tasks. It can be relevant if we want to robustify all the tasks of the model together. This study differs from the threat model and objectives of our paper. Nevertheless, we evaluate in Table \ref{table:atta_both} the three cases. 

Our findings indicate that there is no transferability from the Jigsaw task (the robustness improved from 0 to 1.64\%), weak transferability from the Rotation task which improves the robustness to 12.22\%), and strong transferability from the Macro task (where the robustness improved to 41.98\%). We can explain this phenomenon by how related are the auxiliary tasks to the main task. Indeed, the Macro task is the most related to the main task. We acknowledge that further research is needed in this area, and we appreciate your interest in our work.

Overall, our results suggest that robustifying both tasks require additional optimizations. Indeed, it suggests that $\delta$ generated on both tasks does not robustify the main task for Jigsaw and Rotation, and can significantly deteriorate its clean performance.

\begin{table}[]
\caption{Comparison of GAT for models adversarially trained with a perturbation $\delta$ dependent of the main task only, or a perturbation $\delta$ dependant of both tasks.}  
\label{table:atta_both}
\small
\begin{threeparttable}
\begin{tabular}{l||l|rr}
\hline
      \multirow{2}{*}{AT on}                          &       Auxiliary task             & Test acc & Robust acc \\ \cline{2-4} 
                                 
                                         & None        & 74.49 \%        & 39.09 \%          \\
                                         \hline
\multirow{3}{*}{Main only}  & Jigsaw   & 43.99 \%        & 32.95 \%            \\
                                         & Rotation & 56.51    \%     & 36.13 \%          \\
                                         & Macro    & 73.70  \%       & 48.38    \%       \\
                                         \hline
\multirow{3}{*}{Both tasks} & Jigsaw   & 32.20     \%    & 17.20  \%         \\

                                         & Rotation & 44.90 \%        & 24.80 \%          \\
                                         & Macro    & 75.20 \%        & 36.50 \%     \\
\hline
\multirow{3}{*}{Auxiliary only} & Jigsaw   &    79.70 \%    &   1.64\%         \\

                                         & Rotation &   86.03\%       &  12.22\%          \\
                                         & Macro    &  83.98\%        & 41.98 \%     \\
\hline
\bottomrule
\end{tabular}
\end{threeparttable}
\end{table}

\paragraph{Impact of the number of tasks}

\begin{figure}[t]
\begin{center}
\centerline{\includegraphics[width=\columnwidth]{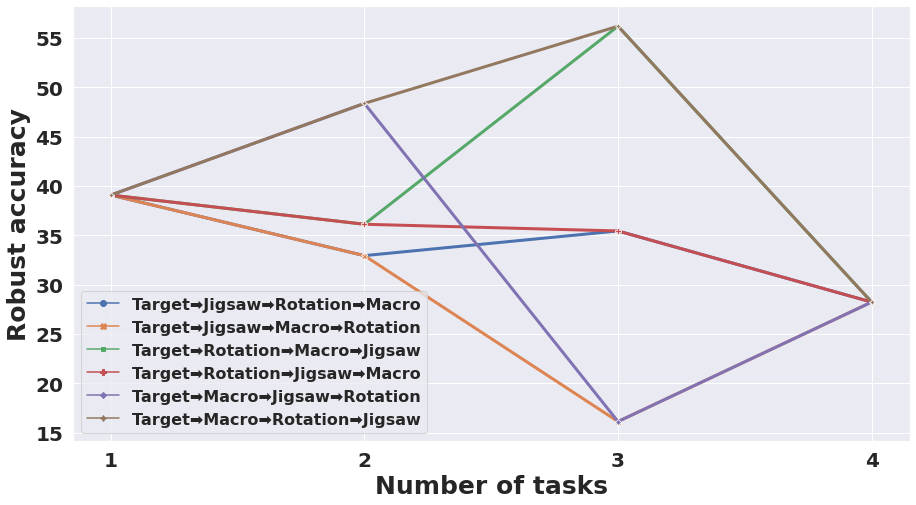}}
\caption{Adversarial robustness' change when adding additional tasks then AT from scratch with GAT.}
\label{fig:all_tasks}
\end{center}
\end{figure}

While our approach is proven for two tasks (because of Theorem \ref{theorem3} and MGDA), we hypothesize that it can still achieve higher robustness than SoTA with additional tasks. We show in Figure \ref{fig:all_tasks} how GAT behaves with additional tasks. Performances peak at three tasks using the combination of \textbf{Macro} and \textbf{Rotation} auxiliary tasks, 
while AT with combinations of three or four tasks involving the \textbf{Jigsaw} task remains less effective.



\section{Generalization studies}
\label{sec:generalization}

\begin{table}[]
\begin{center}
\small
\caption{Robust accuracy of different models AT with GAT, with 3 different task augmentations, compared to their counterpart single task AT. In bold the cases where GAT outperforms single-task AT.
} 
\label{table:generalization}
\small
\begin{threeparttable}
\begin{tabular}{l|rrrr}
\hline
\multirow{2}{*}{Scenario} & \multicolumn{4}{c}{Auxiliary task} \\
 &  \textit{None} &  \textit{Jigsaw} &   \textit{Macro} & \textit{Rotation} \\
\hline
AutoAttack & 27.01 & \textbf{29.63}  &  \textbf{32.54} &    13.82 \\
AutoPGD & 40.00 & 23.00  &  \textbf{40.90} &    21.00 \\
FAB & 61.90 & 34.30  &  \textbf{63.40} &    56.10 \\
\cline{1-5}
Transfer & 1.53 & \textbf{13.44} & \textbf{10.8} & \textbf{15.45} \\
\cline{1-5}
WideResnet28& 42.52 &  32.75 & \textbf{46.6} & 41.06 \\

\hline
\bottomrule
\end{tabular}

\end{threeparttable}
\end{center}
\end{table}

For a fair computational cost comparison, we compared GAT with AT techniques on the same Resnet50 architecture and training protocol. Some AT leverage larger models or datasets, or longer training to achieve better robustness on standardized benchmarks \cite{croce2020robustbench}. We study in the following whether GAT can generalize to more complex datasets and tasks, larger models, or different threat models.


\paragraph{Generalization to dense tasks.}

While our study focused on classification tasks, GAT can be deployed with self-supervised dense tasks like auto-encoders or depth estimation. To confirm the generalization of our approach, we evaluate an additional dataset: \emph{ROBIN}; one of the most recent benchmarks to evaluate robustness of models~\cite{zhao2021robin}. 
We evaluate 3 additional self-supervised dense tasks: Depth estimation, Histogram of Oriented Gradients, and Auto-encoder. We evaluate these tasks both on CIFAR-10 and ROBIN. Our results in Table \ref{table:atta_dense} suggest that dense tasks can also be used to improve the robustness of models.


\begin{table}[]
\caption{Comparison of the performance GAT for models adversarially trained with dense tasks.}  
\label{table:atta_dense}
\small
\begin{threeparttable}
\begin{tabular}{l|l|rr}
\hline
Dataset & Auxiliary Task   & Test acc & Robust acc \\
\hline
\multirow{4}{*}{CIFAR10} & No auxiliary     & 74.49 \%        & 39.09  \%         \\
        & Depth estimation & 43.12   \%      & 0.5  \%           \\
        & HOG              & 83.39  \%       & 44.59 \%          \\
        & Auto-encoder     & 85.11 \%        & 42.23   \%        \\
\hline
\multirow{4}{*}{ROBIN}   & No auxiliary     & 93.71   \%      & 56.78  \%         \\
        & Depth estimation & 87.43   \%      & 18.54  \%         \\
        & HOG              & 98.59   \%      & 94.93    \%       \\
        & Auto-encoder     & 98.78   \%      & 94.84  \%        \\
\hline
\bottomrule
\end{tabular}
\end{threeparttable}
\end{table}

\paragraph{Generalization to larger architectures.}
We train WideResnet28-10 models with GAT and compare their robust accuracy to a single-task WideResnet28-10 model with AT. \textbf{Macro} increases the single-task model's robust accuracy from 42.52\% to 46.6\%. 

\paragraph{Generalization to adaptive attacks.}
To assess if GAT is not a gradient-obfuscation defense, we evaluate our defended models against a large set of adversarial attacks. AutoPGD\cite{autoattack}, a parameter-free gradient attack, and FAB\cite{croce2020minimally}, a white-box boundary attack. Autoattack~\cite{croce2020reliable} is an ensemble that combines the previous white-box attacks with the black-box attack SquareAttack\cite{andriushchenko2020square} in targeted and untargeted threat models.

The results in Table \ref{table:generalization} show that GAT with \textbf{Jigsaw} or \textbf{Macro} auxiliary tasks provide better robustness to AutoAttack than AT with the target task alone. Thus, we confirm that stronger attacks do not easily overcome the robustness provided by GAT. 

\paragraph{Generalization to transfer attacks.}
We evaluate in Table \ref{table:generalization} the threat model where the attacker has access to the full training set but has no knowledge of the auxiliary tasks leveraged by GAT. Models trained with GAT have slightly different decision boundaries from models with common AT. The success rate of surrogate attacks drops from 98.47\% (i.e., 1.53\% robust accuracy) to 84.55\% when we train the target task with \textbf{Rotation} based GAT.

 \paragraph{Generalization to extremely scarce data.}

We restrict the AT of the models to 10\% of the full CIFAR-10 training dataset and compare the performance of AT with GAT. 
GAT with self-supervised tasks and GAT with domain-knowledge tasks both outperform single-task model AT. In particular, the \textbf{Macro} task augmentation boosts the robust accuracy from 8.37\% to 22.42\%.
The detailed results with 10\%, 25\%, and 50\% of training data are in Appendix \ref{appendixB3}.


\section{Related Work}

\paragraph{Multi-task learning}



\citet{vandenhende2021multi} recently proposed a new taxonomy of MTL approaches.
They organized MTL research around two main questions: (i) which tasks should be learned together and (ii) how we can optimize the learning of multiple tasks. For example, there are multiple grouping and weighting strategies such as 
Gradient Vaccine (GV)~\cite{weight_gv}, 
and Project Conflicting Gradients (PCG)~\cite{weight_pcg} that can significantly impact the training of MT models.

Multi-objective optimization for large number of tasks remains also an open-problem, and the recent work from \citet{taskgrouping} investigated which tasks can be combined to improve clean performance.

Our work explores the orthogonal question of robustness: 
(iii) how can we combine auxiliary tasks with AT to improve the adversarial robustness? 

\paragraph{Self-supervised multi-task learning}

Recent work has evaluated the impact of self-supervised tasks on the robustness.
Klingner~\etal~\cite{klingner2020improved} evaluated how the robustness and performances are impacted by MTL for depth tasks. Their study does not tackle at all adversarial training and focuses on vanilla training.
Hendrycks~\etal~\cite{hendrycks2019using} evaluated the robustness of multi-task models with rotation tasks to PGD attacks. However, they do not take into account the MTL in the adversarial process. PGD is applied summed on both the losses and is therefore used as a single task model.

\paragraph{Adversarial training}

The original AT formulation has been improved, either to balance the trade-off between standard and robust accuracy like TRADES~\cite{zhang2019trades} and FAT~\cite{zhang2020fat}, or to speed up the training \cite{shafahi2019free, wong2020fast}. Finally, AT was combined with data augmentation techniques, either with unlabeled data \cite{carmon2019unlabeled}, self-supervised pre-training \cite{chen2020adversarial}, or Mixup \cite{rebuffi2021data}.

These approaches, while very effective, entail a computation overhead that can be prohibitive for practical cases 
like medical imaging. Our work suggests that GAT is a parallel line of research and can be combined with these augmentations. 

\paragraph{Provable robustness}
This type of robustness is generally not comparable to empirical robustness (which we target) and is not considered in established robustness benchmarks like RobustBench\cite{croce2020robustbench}. Provable robustness of MTL is therefore an orthogonal field to our research.
\section*{Conclusion}

In this paper, we demonstrated that augmenting single-task models with self-supervised and domain-knowledge auxiliary tasks significantly improves the robust accuracy of classification. We proposed a novel adversarial training approach, Guided Adversarial Training  that solves the min-max optimization of adversarial training through the prism of Pareto multi-objective learning and curvature regularization. Our approach complements existing data augmentation techniques for robust learning and improves adversarially trained models' clean and robust accuracy. We expect that combining data augmentation and task augmentation is key for further breakthroughs in adversarial robustness.

\section*{Acknowledgements}
This work is mainly supported by the Luxembourg National Research Funds (FNR) through CORE project C18/IS/12669767/STELLAR/LeTraon

Jingfeng Zhang was supported by JST ACT-X Grant Number JPMJAX21AF and JSPS KAKENHI Grant Number 22K17955, Japan.

\bibliography{general,atks}
\bibliographystyle{icml2023}

\newpage
\appendix
\onecolumn

\section{Appendix A: Replication}

\subsection{Proofs}
\label{appendixA1}

\begin{definition}

Let $\mathscr{M}$ be a multi-task model. $\mathscr{T}' \subseteq \mathscr{T}$ a subset of its tasks and $\mathcal{L}_\mathscr{T}'$ the joint loss of tasks in $\mathscr{T}'$. Then, we call $\mathbb{E}_{x}[\delta\mathcal{L}(\mathscr{T}',\epsilon)]$ the \emph{adversarial vulnerability} of $\mathscr{M}$ on $\mathscr{T}'$ to an $\epsilon$-sized $\| .\|_p$-attack. 

And we define it as the average increase of $\mathcal{L}_{\mathscr{T}'}$ after attack over the whole dataset, i.e.:

\begin{equation*}
    \mathbb{E}_{x}[\delta \mathcal{L}(\mathscr{T}',\epsilon)]=  \mathbb{E}_{x}\left[\max _{\|\delta\|_{p}\leq\epsilon} \mid \mathcal{L}_{\mathscr{T}'}(x+\delta, \bar{y})-\mathcal{L}_{\mathscr{T}'}(x, \bar{y}) \mid \right]
\end{equation*}

\end{definition}

\begin{lemma} 
Under an $\epsilon$-sized $\| .\|_p$-attack, the adversarial vulnerability of a multi-task model can be approximated through the first-order Taylor expansion, that is:
\begin{equation}
    \mathbb{E}_{x}[\delta \mathcal{L}'(x, \bar{y}, \epsilon, \mathscr{T}')]  \propto \mathbb{E}_{x}[\mid\mid \partial_{x} \mathcal{L}'(x, \bar{y}) \mid \mid_q] 
\end{equation}
\label{lemme2}
\end{lemma}

\begin{proof}
\vskip -0.3in

From definition 1, we have:

\begin{multline*}
    \mathbb{E}_{x}[\delta \mathcal{L}(\mathscr{T}',\epsilon)]=  \mathbb{E}_{x}\left[\max _{\|\delta\|_{p}\leq\epsilon} \mid \mathcal{L}_{\mathscr{T}'}(x+\delta, \bar{y})-\mathcal{L}_{\mathscr{T}'}(x, \bar{y}) \mid \right]
\end{multline*}

Given the perturbation $\delta$ is minimal, we can approximate $\delta \mathcal{L}$ with a Taylor expansion up to the first order:

\begin{multline*}
    \mathbb{E}_{x}[\delta \mathcal{L}(\mathscr{T}',\epsilon)] =  \mathbb{E}_{x}\left[\max _{\|\delta\|_{p}\leq\epsilon} \mid \mathcal{L}_{\mathscr{T}'}(x+\delta, \bar{y})-\mathcal{L}_{\mathscr{T}'}(x, \bar{y}) \mid \right] \approx  \mathbb{E}_{x}\left[\max _{\|\delta\|_{p}\leq\epsilon} \mid \delta \cdot \partial_{x} \mathcal{L}'(x, \bar{y}) \mid \right]
\end{multline*}

The noise $\delta$ is optimally adjusted to the coordinates of $\partial_{x} \mathcal{L}'$
 within an $\epsilon$-constraint. By the definition of the dual-norm, we get:

\begin{equation}
    \mathbb{E}_{x}[\delta \mathcal{L}'(x, \bar{y}, \delta, \mathscr{T}')]\approx   \mid\mid \delta \mid \mid_p \cdot \mathbb{E}_{x}[\mid\mid 
    \partial_{x} \mathcal{L}'(x, \bar{y})\mid \mid_q]
\end{equation}

where $q$ is the dual norm of $p$ and $\frac{1}{p}+\frac{1}{q}=1$ and $1 \leq p \leq \infty$.

Once given the p-norm bounded ball, i.e., $\mid\mid \delta \mid \mid_p$ is constant (denoted $C$ in the following), we get:

\begin{equation}
    \mathbb{E}_{x}[\delta \mathcal{L}'(x, \bar{y}, \epsilon, \mathscr{T}')]\approx C \cdot \mathbb{E}_{x}[\mid\mid 
     \partial_{x} \mathcal{L}'(x, \bar{y})
    \mid \mid_q] \propto \mathbb{E}_{x}[\mid\mid \partial_{x} \mathcal{L}'(x, \bar{y}) \mid \mid_q] 
\end{equation}

\end{proof}

\begin{theorem}

Consider a multi-task model $\mathscr{M}$ where an attacker targets $\mathscr{T}= \{t_1,t_2\}$ two tasks weighted with $\alpha_1$ and $\alpha_2$ respectively, with an $\epsilon$-sized $\| .\|_p$-attack. If the model is converged, and the gradient for each task is i.i.d. with zero mean and the tasks are correlated, the adversarial vulnerability of the model can be approximated as

\begin{equation}
\begin{aligned}
    \mathbb{E}_{x}[\delta \mathcal{L}'] \propto \sqrt{1+ 2 \frac{\alpha_1 . \alpha_2 . \operatorname{Cov}\left(\partial_{x} \mathcal{L}_1, \partial_{x} \mathcal{L}_2\right)}{\alpha_1^2 \sigma_1^2+ \alpha_2^2 \sigma_2^2 }},
\end{aligned}
\end{equation}
\label{theorem3_appendix}
\end{theorem} 

where $ \sigma_i^2 = \operatorname{Cov}\left(\partial_{x} \mathcal{L}_i, \partial_{x} \mathcal{L}_i\right)$ and $\partial_{x} \mathcal{L}(x, y_i)$ the gradient of the task $i$.

\subsection{}
\begin{proof}

let $\mathbf{r}_{i}=\alpha_i \cdot \partial_{x} \mathcal{L}(x, y_i)$ the weighted gradient of the task i, with a weight $\alpha_i$ such as the joint gradient of $\mathscr{M}$ is defined as $\partial_{x} \mathcal{L}(x, \bar{y}) = \sum_{i=1}^{M}\mathbf{r}_{i}$.  let $p=q=2$

We have:

\begin{equation}
\begin{aligned}
   \mathbb{E}_{x}\left[\mid\mid 
    C \cdot \partial_{x} \mathcal{L}'(x, \bar{y})
    \mid \mid_2^2\right] = \mathbb{E}_{x}\left[\mid\mid 
    \sum_{j=1}^{M} C \cdot r_i
    \mid \mid_2^2\right] \\
    =C^2 \mathbb{E}_{x}\left[
    \sum_{i=1}^{M} \mid\mid  \mathbf{r}_{i} \mid\mid_2^2 + 
    2\sum_{i=1}^{M}\sum_{j=1}^{i-1} \mid\mid\mathbf{r}_{i}\mid\mid_2\mid\mid\mathbf{r}_{j} \mid\mid_2
    \right]\\
    =C^2 \left( \sum_{i=1}^{M} \mathbb{E}_{x}[\mathbf{r}_{i}^2] + 2\sum_{i=1}^{M}\sum_{j=1}^{i-1} \mathbb{E}_{x}[\mathbf{r}_{i}\mathbf{r}_{j}] \right)
\end{aligned}
\end{equation}

For two tasks, we have then:

\begin{equation}
\begin{aligned}
   \mathbb{E}_{x}\left[\mid\mid 
    C \cdot \partial_{x} \mathcal{L}'(x, \bar{y})
    \mid \mid_2^2\right]
    =C^2 \left( \mathbb{E}_{x}[\mathbf{r}_{1}^2] + \mathbb{E}_{x}[\mathbf{r}_{2}^2] + 2  \mathbb{E}_{x}[\mathbf{r}_{1}\mathbf{r}_{2}] \right)
\end{aligned}
\end{equation}

We know:
\begin{equation}
\begin{aligned}
\operatorname{Cov}\left(\mathbf{r}_{i}, \mathbf{r}_{j}\right) =  \mathbb{E}_{x}\left[\mathbf{r}_{i}\mathbf{r}_{j}\right]- \mathbb{E}_{x}\left[\mathbf{r}_{i}\right]\mathbb{E}_{x}\left[\mathbf{r}_{j}\right]
\end{aligned}
\end{equation}

According to the assumptions, the gradient of each task is i.i.d with zero means: $\mathbb{E}_{x}\left[\mathbf{r}_{i}\right]=0$
Then $\operatorname{Cov}\left(\mathbf{r}_{i}, \mathbf{r}_{j}\right) = \mathbb{E}_{x}\left[\mathbf{r}_{i}\mathbf{r}_{j}\right]$ and
$\mathbb{E}_{x}\left[\mathbf{r}_{i}^2\right] = \operatorname{Cov}\left(\mathbf{r}_{i}, \mathbf{r}_{i}\right)  = \alpha_i^2 \operatorname{Cov}\left(\partial_{x} \mathcal{L}_i, \partial_{x} \mathcal{L}_i\right) = \alpha_i^2 \sigma_i^2 $.

\begin{equation}
\begin{aligned}
   \mathbb{E}_{x}\left[\mid\mid 
    C \cdot \partial_{x} \mathcal{L}'(x, \bar{y})
    \mid \mid_2^2\right] = 
    {C^2} \left(\alpha_1^2 \sigma_1^2 + \alpha_2^2 \sigma_2^2 + 2 \cdot \alpha_1 \alpha_2 \operatorname{Cov}\left(\partial_{x} \mathcal{L}_1, \partial_{x} \mathcal{L}_2\right)
    \right) \\
    = K \cdot 
    \left(1 + 2\frac{\alpha_1 \alpha_2 \operatorname{Cov}\left(\partial_{x} \mathcal{L}_1, \partial_{x} \mathcal{L}_2\right)}{\alpha_1^2 \sigma_1^2 + \alpha_2^2 \sigma_2^2}
    \right) \\
     C \cdot \mathbb{E}_{x}\left[\mid\mid 
    \partial_{x} \mathcal{L}'(x, \bar{y})
    \mid \mid_2\right] = \sqrt{K} \cdot  \sqrt{\left(1 + 2\frac{\alpha_1 \alpha_2 \operatorname{Cov}\left(\partial_{x} \mathcal{L}_1, \partial_{x} \mathcal{L}_2\right)}{\alpha_1^2 \sigma_1^2 + \alpha_2^2 \sigma_2^2}
    \right)}
\end{aligned}
\end{equation}

where $K=C^2 (\alpha_1^2 \sigma_1^2 + \alpha_2^2 \sigma_2^2) $

Using the first order adversarial vulnerability (Lemma 2), we then have:

\begin{equation}
\begin{aligned}
    \mathbb{E}_{x}[\delta \mathcal{L}']\approx  C \cdot \mathbb{E}_{x}\left[\mid\mid 
    \partial_{x} \mathcal{L}'(x, \bar{y})
    \mid \mid_2\right] \approx \sqrt{K} \cdot \sqrt{1 + 2\frac{\alpha_1 \alpha_2 \operatorname{Cov} \left(\partial_{x} \mathcal{L}_1, \partial_{x} \mathcal{L}_2 \right)}{\alpha_1^2 \sigma_1^2 + \alpha_2^2 \sigma_2^2}}
     \propto \sqrt{1+ 2 \frac{\alpha_1 . \alpha_2 . \operatorname{Cov}\left(\partial_{x} \mathcal{L}_1, \partial_{x} \mathcal{L}_2\right)}{\alpha_1^2 \sigma_1^2+ \alpha_2^2 \sigma_2^2 }}
\end{aligned}
\end{equation}

with K a constant dependent of the bounded ball and the attacked tasks.

\end{proof}
\newpage
\subsection{GAT Algorithm}
\label{appendixA2}

\begin{algorithm}[t]
\caption{Pseudo-Algorithm of GAT}
\begin{algorithmic}

\STATE{\textbf{Given}}: a single task model $\mathscr{M}$ parameterized by $\theta^{s}$ for the shared encoder and $\theta^{t}$ for the specific heads, a batch example $x$, and $\bar{y}= (y_1,..., y_s, ... y_m)$ its corresponding labels for each task, with $y_1$ the target task, $y_{1 < i \leq s}$ the auxiliary self-supervised tasks and $y_{s < i \leq m}$ the auxiliary domain-knowledge tasks; 
\STATE{\textbf{Given}}: an input processing $f_t$ for each auxiliary self-supervised $t$ task with label $y_{1 < t \leq s}$.
\STATE{\textbf{Given}}: a weight optimizer $opt$; a list of task-specific decoders functions $\mathscr{D} = \{D_1,...,D_M\}$
\STATE{\textbf{Given}}: a $PGD$ adversarial attack with a step size $\epsilon_{step}$; a maximum perturbation $\epsilon$; $S$ number of attack iterations;

 \STATE{\textbf{Step 1}:} Create a decoder $D_i$ at the penultimate layer of $\mathscr{M}$ for each of the auxiliary task $t_i$ / $i>1$. 
       
\STATE{\textbf{Step 2}:} For each epoch and batch $x$ 
Do
       \begin{enumerate}

           \item For each self-supervised task $t_{1 < i \leq s}$, successively pre-process the batch examples $x$ with the appropriate input processing function:  \newline 
           $x \leftarrow  \bigcircop_{t=2}^s \ f_{t}(x)$
           \item Generate $\hat{x}$, the adversarial examples of $x$: $\hat{x} \leftarrow \mathrm{PGD}(x, y_1, \epsilon_{step}, \epsilon, S)$.
           \item Compute the losses $l_{i,x}$ and $l_{i,\hat{x}}$ of $x$ and $\hat{x}$ respectively for each task $t_i$ with label $y_i$; $l \leftarrow l_{1,x},l_{1,\hat{x}},\ldots , l_{M,x},l_{M,\hat{x}}, l^{(reg)}_{M+1,x} ,\ldots , l^{(reg)}_{2M-1,x}$.
           \item Apply MGDA to find the minimum norm element in the convex hull given the list of losses:  \newline
           $\alpha_1,\ldots,\alpha_{2M-1} \leftarrow$ MGDA($\theta^{s},\theta^t\theta$,$l$)
           \item Back-propagate the weighted gradients and update the model weights with optimizer $opt$.
           \newline 
           $\theta^{s} \leftarrow \theta^{s} -   \eta \sum_{t=1}^{2M-1} \alpha_{t} \nabla_{\theta^{sh}}  l_{t,x}(\theta^{s},\theta^t)$
       \end{enumerate}
\STATE{\textbf{Step 3}:} Disable the auxiliary branches added at step 1.
\end{algorithmic}
\vspace{-1mm}
\end{algorithm}

Following \citet{weight_mgda}, we use the Frank-Wolfe algorithm\cite{frank-wolfe} to solve the constrained optimization problem as follows:

\begin{algorithm}[]
\caption{MGDA($\theta^{s},\theta^t$,$l$) procedure \cite{weight_mgda}}
\label{alg:mtl_mgda_solver}
\begin{algorithmic}

\STATE{\textbf{Initialize}} $\alpha = (\alpha^1, \ldots, \alpha^{T}) = (\frac{1}{T}, \ldots, \frac{1}{T})$

\STATE{\textbf{Precompute $\mathcal{M}$}} st. $\mathcal{M}_{i,j} = \big(\nabla_{\theta^{sh}}  \hat{l}_i(\theta^{sh},\theta^i)\big)^\intercal \big(\nabla_{\theta^{sh}}  \hat{l}_j(\theta^{sh},\theta^j)\big)$

\STATE{\textbf{Repeat}}

\STATE{} $\hat{t} = \argmin_r \sum_t \alpha^t \mathcal{M}_{rt}$

\STATE{} $\hat{\gamma} = \argmin_{\gamma}  \big( (1 - \gamma) \bm{\alpha} + \gamma \bm{e}_{\hat{t}}  \big)^\intercal \mathcal{M}  \big( (1 - \gamma) \bm{\alpha} + \gamma \bm{e}_{\hat{t}}  \big)$

\STATE$\bm{\alpha} = (1- \hat{\gamma})\bm{\alpha} + \hat{\gamma} \bm{e}_{\hat{t}}$

\STATE{\textbf{until}} $\hat{\gamma} \sim 0$ {\bfseries or} Number of Iterations Limit
\STATE{\textbf{return}} $\alpha^1,\ldots,\alpha^{T}$

\end{algorithmic}
\end{algorithm}

\subsection{Experimental Setting}
\label{appendixSetup}

\subsubsection{Datasets}
\label{appendixA3}

We show in table \ref{tab:xrays_meta} the general properties of the datasets used in training our models. Table \ref{tab:path_counts} (\cite{cohen2020limits}) details the number of positive and negative examples with each label for each dataset. Our models are trained either on CheXpert or NIH depending on the evaluation. 

Our evaluation covers as target tasks very scarce pathologies (Edema, Pneumonia), and medium scarce pathologies (Atelectasis), across both datasets.

\begin{table*}[h]
    \centering \begin{tabular}{l|ll}
\toprule
& NIH & CheXpert \\ 
\midrule
Number of patient radiographs & 112,120                   & 224,316                   \\
Number of patients            & 30,805                    & 65,240                    \\
Age in years: mean (standard deviation)        & 46.9 (16.6)               & 60.7 (18.4)               \\
Percentage of females (\%)    & 43.5\%           		  & 40.6\%           \\
Number of pathology labels    & 8                        & 14                        \\
\bottomrule
\end{tabular}

\caption{Characteristics of NIH and CheXpert datasets used in our evaluation.}
\label{tab:xrays_meta}
\end{table*}

\begin{table*}[h]
\centering
\begin{tabular}{ccc}
\hline
Dataset                & NIH        & CheXpert    \\ \hline
\textbf{Atelectasis}   & 1702/29103 & 12691/14317 \\
\underline{Cardiomegaly}  & 767/30038  & 9099/17765  \\
Consolidation & 427/30378  & 5390/22504  \\
\textbf{Edema}         & 82/30723   & 14929/20615 \\
Effusion      & 1280/29525 & 20640/23500 \\
Emphysema              & 265/30540  & -           \\
Enlarged Cardio        & -          & 5181/20506  \\
Fibrosis               & 571/30234  & -           \\
Fracture               & -          & 4250/14948  \\
Hernia                 & 83/30722   & -           \\
Infiltration           & 3604/27201 & -           \\
Lung Lesion            & -          & 4217/14422  \\
Lung Opacity           & -          & 30873/15675 \\
Mass                   & 1280/29525 & -           \\
Nodule                 & 1661/29144 & -           \\
Pleural Thickening     & 763/30042  & -           \\
\textbf{Pneumonia}     & 168/30637  & 2822/14793  \\
\underline{Pneumothorax}  & 269/30536  & 4311/32685  \\ \hline
\end{tabular}

    \caption{Samples distributions across each pathology and dataset. Each cell shows the number of positive/negative samples of the label. There are 7 common pathologies in NIH and CheXpert datasets. Among those, in bold the pathologies evaluated as target task, and in underline the pathologies used as an auxiliary. }
    \label{tab:path_counts}
\end{table*}

All datasets, CheXpert, NIH, and ROBIN use images of the same dimensions as ImageNet (256x256). We did not include Tiny ImageNet in our study as it lacks multiple tasks required for a comprehensive evaluation. Furthermore, we chose the ROBIN dataset as it is a subset of ImageNet that offers additional labels that can serve as auxiliary task, making it the most suitable variant for our study.

\subsection{Architectures}

The majority of the tests are carried out using the Resnet50v2 \cite{he2016resnet50} encoder, which has a depth of 50 and 25.6M parameters. This encoder is the main focus because it is the most widely used for Xray image classification \cite{ganesan2019assessment}.
We also perform some tests using the WRN-28-10 \cite{zagoruyko2016wide} encoder, which has a depth of 28, a width multiplier of 10, and 36M parameters.

\subsection{Adversarial Training}

\paragraph{The outer minimization:} We use MADRY adversarial training \cite{madry2017towards}, i.e. we train the model using a summed loss computed from the clean and adversarial examples. for ATTA, we use a backpropagation over the pareto optimal of the four losses. The learning uses the SGD optimizer with lr=0.1, a cosine annealing, and checkpoint over the best performance. 

\paragraph{The inner maximization:} We generate the adversarial examples with PGD \cite{pgd_madry2019deep}, on $\ell_\infty$ norms and $\epsilon=8/255$ for CIFAR-10 and STL-10 and $\epsilon=4/255$ for CheXpert and NIH models. We use in the iterative attack 1 random start, and 10 steps.

\subsection{Robustness evaluation}

We evaluate the robustness against PGD-10 on $\ell_\infty$ norms and $\epsilon=8/255$ for CIFAR-10 and STL-10 and $\epsilon=4/255$ for CheXpert and NIH models.
We also evaluate CIFAR-10 models against AutoAttack \cite{autoattack}. Autoattack is a mixture of $\ell_\infty epsilon=8/255$ attacks: untargeted AUTOPGD (a variant of PGD with an adaptive step) on the cross-entropy loss with 100 steps, targeted AUTOPGD with 100 steps, a 100 steps FAB attack, and finally a 5000 queries Square attack.

These hyper-parameters of AutoAttack are consistent with AutoAttack's default parameterization in \citet{kim2020torchattacks, croce2020robustbench}.

\subsection{Computation budget}

We train all our models on slurm nodes, using single node training. Each node has one A100 GPU 32Gb V100 SXM2. 
We train CIFAR-10 and STL-10 models for 400 epochs and CheXpert and NIH models for 200 epochs. The WRN-70-16 model is trained for 40 epochs to account for being 10 times larger than the Resnet50 used for the main evaluation.

Our license is \textbf{MIT Licence}, and we use the following external packages:

\paragraph{Torchxrayvision:} Located in folder \textbf{./torchxrayvision}. Adapted from \url{https://github.com/mlmed/torchxrayvision}: Apache Licence 
\paragraph{Taskonomy/Taskgrouping:} Located in folder \textbf{./utils/multitask\_models}. Adapted from \url{https://github.com/tstandley/taskgrouping/} MIT Licence
\paragraph{LibMTL:} Located in folder \textbf{./utils/weights}. Adapted from \url{https://github.com/median-research-group/LibMTL} MIT Licence

\clearpage
\newpage

\section{Appendix B: Detailed results of the main study}

\subsection{Limited data training with CIFAR-10}
\label{appendixB1}

\paragraph{GAT:}
To evaluate whether test accuracy (i.e. equal weights task augmentation) is effective when access to adversarial training data is limited, we first train models with the full dataset for 200 epochs then we adversarial fine-tune (PGD-4; 8/255) the models with a subset of training data (10\%, 50\%).  For each scenario, we fine-tune 3 different models with different seeds and report in Table \ref{table:partial_adv} the Test Accuracy (Test accuracy) and Robust Accuracy (Robust accuracy) with and without an auxiliary task. We report the mean and standard deviation across the runs. The std across the experiments is pretty low and the conclusions of the main paper hold.

\begin{table*}[]
\begin{center}
\caption{Evaluation results of 4 Different ($\mathcal{D}_i, \mathcal{T}_i, \mathcal{A}_i$) Scenarios: $\mathcal{D}_1$ (adversarial fine-tuning with 10\% of the training data), $\mathcal{D}_2$ (adversarial fine-tuning with 50\% of the training data), $\mathcal{T}_{1,2,3}$ training respectively without an auxiliary task, with Rotation and with Jigsaw task, $\mathcal{A}_1$ (Robust Accuracy against a PGD-4 attack), $\mathcal{A}_2$ (Robust Accuracy against a PGD-10 attack).} 
\label{table:partial_adv}
\begin{threeparttable}

\begin{tabular}{llllrr}
\toprule

Dataset subset & Auxiliary & PGD steps & Metric &       mean &   std     \\
\midrule
0.1 & \textit{None} & 10 & Test accuracy &  60.41 &  0.62 \\
    &   & & Robust accuracy &   8.37 &  0.32 \\
    & & 4 & Test accuracy &  60.38 &  0.59 \\
    & &  & Robust accuracy &  11.81 &  0.23 \\
        & \textit{Jigsaw} & 10 & Test accuracy &  51.98 &  0.53 \\
    &         &   & Robust accuracy &  32.41 &  0.46 \\
    &         & 4 & Test accuracy &  51.06 &  1.47 \\
    &        &   & Robust accuracy &  32.26 &  0.85 \\
 & \textit{Rotation} & 10 & Test accuracy &  50.41 &  0.11 \\
    &          &   & Robust accuracy &  15.01 &  0.29 \\
    &          & 4 & Test accuracy &  50.17 &  0.49 \\
    &          &   & Robust accuracy &  20.01 &  0.27 \\
    & \textit{Macro} & 10 & Test accuracy &  65.62 & 0.48  \\
    &         &   & Robust accuracy &  22.42 &  0.35 \\
    &         & 4 & Test accuracy &  65.65 &   0.42 \\
    &        &   & Robust accuracy &  42.68 & 0.38 \\
\midrule
0.5 & \textit{None} & 10 & Test accuracy &  77.45 &  0.25 \\
    & &   & Robust accuracy &  25.04 &  0.15 \\
    & & 4 & Test accuracy &  77.51 &  0.15 \\
    & &   & Robust accuracy &  31.71 &  0.08 \\
    & \textit{Jigsaw} & 10 & Test accuracy &  59.72 &  0.45 \\
    &          &   & Robust accuracy &  29.08 &  1.58 \\
    &          & 4 & Test accuracy &  58.42 &  1.16 \\
    &          &   & Robust accuracy &  33.68 &  0.52 \\
 & \textit{Rotation} & 10 & Test accuracy &  59.77 &  0.58 \\
    &          &   & Robust accuracy &  17.09 &  0.51 \\
    &          & 4 & Test accuracy &  59.69 &  0.70 \\
    &          &   & Robust accuracy &  24.56 &  0.49 \\
    & \textit{Macro} & 10 & Test accuracy & 73.68  & 0.72  \\
    &          &   & Robust accuracy & 33.76  &  0.71 \\
    &          & 4 & Test accuracy & 73.62  &   0.81\\
    &          &   & Robust accuracy & 54.14  &   0.63\\
\bottomrule
\end{tabular}

\end{threeparttable}
\end{center}
\vspace{-3mm}
\end{table*}


\subsection{CheXpert detailed results}
\label{appendixB2}

We extend the evaluation of the main paper to 6 additional combinations of auxiliary tasks and target task, using the \textbf{Pneumonia} pathology as a target. We present all the results in Table \ref{table:chexpert_adv}. These extended results corroborate that AT with auxiliary task significantly improves the robustness of classification models on the CheXpert dataset \cite{irvin2019chexpert}.

\begin{table*}[]
\begin{center}
\caption{GAT different data scenarios: 10\%, 25\% and 50\% of CIFAR-10 dataset. We evaluate 3 different task augmentations with MGDA weighting strategy.
} 
\label{table:full_adv_partial}
\begin{threeparttable}
\begin{tabular}{l|r|r|r|r|r|r|r|r}
\hline
Scenario  & \multicolumn{4}{c|}{Clean accuracy (\%)} & \multicolumn{4}{c}{Robust accuracy (\%)} \\
\hline
 &  \textit{None} & \textit{Jigsaw} &   \textit{Macro} & \textit{Rotation} &  \textit{None} &   \textit{Jigsaw} &   \textit{Macro} & \textit{Rotation} \\

10\%   & 52.66 & 42.7 &  54.89 &    47.07 &  12.46 & 32.14 & 13.43 &   39.2 \\

25\% & 68.39 & 49.76& 68.54 &  62.85   & 24.56 &32.08  &  27.74 &    47.75 \\

50\% & 76.13 &  66.57 &  76.5 &  78.19   & 33.69 & 23.79  &  31.94 &   16.63 \\

\bottomrule
\end{tabular}

\end{threeparttable}
\end{center}
\vspace{-3mm}
\end{table*}

\begin{table}[]
\begin{center}
\caption{Robust and clean AUC of CheXpert models trained with GAT.} 
\label{table:chexpert_adv}
\begin{threeparttable}
\begin{tabular}{llrr}
\toprule
 Target Task & Auxiliary Task &  Robust AUC &  Clean AUC \\
\midrule
 Atelectasis &    Single task &            50.00 &           58.76 \\
 Atelectasis &   Cardiomegaly &            71.20 &           71.97 \\
 Atelectasis &   Pneumothorax &            70.93 &           71.92 \\
 Atelectasis &            Age &            46.26 &           66.89 \\
 Atelectasis &         Gender &            83.00 &           83.35 \\
 Atelectasis &         Jigsaw &            63.81 &           65.92 \\
 Atelectasis &       Rotation &            78.32 &           74.50 \\
       Edema &    Single task &            55.69 &           52.42 \\
       Edema &   Cardiomegaly &            52.74 &           55.79 \\
       Edema &   Pneumothorax &            47.40 &           58.86 \\
       Edema &            Age &            59.17 &           53.41 \\
       Edema &         Gender &            31.46 &           56.07 \\
       Edema &         Jigsaw &            70.47 &           67.77 \\
       Edema &       Rotation &            52.59 &           55.98 \\
   Pneumonia &    Single task &            38.70 &           56.66 \\
   Pneumonia &   Cardiomegaly &            57.47 &           57.05 \\
   Pneumonia &   Pneumothorax &            32.25 &           56.74 \\
   Pneumonia &            Age &            49.15 &           56.58 \\
   Pneumonia &         Gender &            60.08 &           57.59 \\
   Pneumonia &         Jigsaw &            46.45 &           56.47 \\
   Pneumonia &       Rotation &            60.76 &           60.00 \\
\bottomrule
\end{tabular}

\end{threeparttable}
\end{center}
\vspace{-3mm}
\end{table} 

\subsection{CIFAR10 detailed results}
\label{appendixB3}

\begin{table}[]
\small
\caption{Robust and clean Accuracy of CIFAR-10 models trained with GAT vs trained with other adversarial training (AT) optimizations}
\begin{threeparttable}
\begin{tabular}{lrrl}

\hline
Method                              & Robust accuracy (\%) & Clean accuracy (\%) & Type of AT optimization                 \\ \hline
\textbf{GAT} [Ours]                          & 48.38               & 73.70              &      Task augmentation                 \\
GAT-noReg [Ours]                    & 44.34               & 83.47              &                Task augmentation       \\
CutMix \cite{yun2019cutmix}                & 38.95               & 87.31              & Data augmentation     \\
Maxup \cite{maxup}                & 45.24               & 83.54              & Data augmentation     \\
Unlabeled \cite{carmon2019unlabeled}         & 21.98               & 80.30              & Data augmentation     \\
DDPM \cite{gowal2021improving}              & 44.41               & 73.27              & Data augmentation \\
TRADES \cite{zhang2019trades}              & 42.76               & 60.73              & Training optimization \\
FAST \cite{wong2020fast}                 & 26.56               & 78.18              & Training optimization \\
Self-supervised \cite{hendrycks2019using}  & 36.50               & 75.20              & Training \& data optimization \\
Pre-training \cite{chen2020adversarial}          & 27.30               & 86.64              & Training \& data optimization \\
Madry Adversarial Training \cite{pgd_madry2019deep} & 39.09               & 74.49              &    Training optimization                   \\ \hline
\end{tabular}
\end{threeparttable}
\label{table:sota_cifar10_supp}
\end{table}

We gather the performance of all SoTA adversarial training approaches in Table \ref{table:sota_cifar10_supp}.
GAT without regularization term has a lower robustness but preserves better the clean performance.
For DDPM, we follow \citet{gowal2021improving} and use samples generated by a Denoising Diffusion Probabilistic Model \cite{ddpm} to improve robustness. The DDPM is solely trained on the original training data and does not use additional external data. We do not however use the additional optimizations proposed by \citet{gowal2021improving} to achieve their results, and stick to the same training protocol as all our experiments.

Their additional training optimizations are detailed in their repository: \url{https://github.com/deepmind/deepmind-research/tree/master/adversarial_robustness/pytorch}.
\clearpage
\newpage
\section{Appendix C: Complementary results}
\label{appendixComplementary}
\subsection{Statistical significance when compared with SoTA}

Dietterich suggests the McNemar's test in seminal study on the use of statistical hypothesis tests to compare classifiers\cite{dietterich}.

The test is particularly suggested when the methods being compared can only be assessed once, e.g. on a single test set, as opposed to numerous evaluations using a resampling methodology, such as k-fold cross-validation. It is also recommended when the computation cost of training the same model multiple times is high. Both are our cases in this study.

The results in Table 2 of the main paper use this statistical test. The blue cells are the ones where we can reject Null Hypothesis: Classifiers with ATTA vs without have a different proportion of errors on the test set. In our study, we use $\alpha = 0.05$ and provide in the figshare repository \url{https://figshare.com/projects/ATTA/139864} the \emph{Contingency tables} and raw values of the test: You can find the sumary of the evaluation below:

\begin{multicols}{3}
\small

\begin{itemize}
    \item Model 'Rotation' VS 'Macro'

 statistic=792.000, p-value=0.000

 Different proportions of errors (reject H0): \textbf{The two classifiers have a different proportion of errors on the test}
\item Model 'Rotation' VS 'Jigsaw'

 statistic=610.000, p-value=0.000

 reject H0
\item Model 'Rotation' VS 'Depth'

 statistic=464.000, p-value=0.000

 reject H0
\item Model 'Rotation' VS 'Hog'

 statistic=770.000, p-value=0.000

 reject H0
\item Model 'Rotation' VS 'Rot + Unlabeled'

 statistic=395.000, p-value=0.000

 reject H0
\item Model 'Rotation' VS 'Macro + Unlabeled'

 statistic=380.000, p-value=0.000

 reject H0
\item Model 'Rotation' VS 'Unlabeled'

 statistic=788.000, p-value=0.000

 reject H0
\item Model 'Rotation' VS 'Jigsaw + Unlabeled'

 statistic=638.000, p-value=0.000

 reject H0
\item Model 'Rotation' VS 'Rot + Cutmix'

 statistic=610.000, p-value=0.000

 reject H0
\item Model 'Rotation' VS 'Macro + Cutmix'

 statistic=842.000, p-value=0.010

 reject H0
\item Model 'Rotation' VS 'Jigsaw + Cutmix'

 statistic=590.000, p-value=0.000

 reject H0
\item Model 'Macro' VS 'Rotation'

 statistic=792.000, p-value=0.000

 reject H0
\item Model 'Macro' VS 'Jigsaw'

 statistic=502.000, p-value=0.000

 reject H0
\item Model 'Macro' VS 'Depth'

 statistic=372.000, p-value=0.000

 reject H0
\item Model 'Macro' VS 'Hog'

 statistic=460.000, p-value=0.095

 Same proportions of errors (fail to reject H0)
\item Model 'Macro' VS 'Rot + Unlabeled'

 statistic=509.000, p-value=0.000

 reject H0
\item Model 'Macro' VS 'Macro + Unlabeled'

 statistic=287.000, p-value=0.000

 reject H0
\item Model 'Macro' VS 'Unlabeled'

 statistic=715.000, p-value=0.103

 Same proportions of errors (fail to reject H0)
\item Model 'Macro' VS 'Jigsaw + Unlabeled'

 statistic=472.000, p-value=0.000

 reject H0
\item Model 'Macro' VS 'Rot + Cutmix'

 statistic=621.000, p-value=0.000

 reject H0
\item Model 'Macro' VS 'Macro + Cutmix'

 statistic=631.000, p-value=0.010

 reject H0
\item Model 'Macro' VS 'Jigsaw + Cutmix'

 statistic=468.000, p-value=0.000

 reject H0
\item Model 'Jigsaw' VS 'Rotation'

 statistic=610.000, p-value=0.000

 reject H0
\item Model 'Jigsaw' VS 'Macro'

 statistic=502.000, p-value=0.000

 reject H0
\item Model 'Jigsaw' VS 'Depth'

 statistic=1066.000, p-value=0.000

 reject H0
\item Model 'Jigsaw' VS 'Hog'

 statistic=523.000, p-value=0.000

 reject H0
\item Model 'Jigsaw' VS 'Rot + Unlabeled'

 statistic=344.000, p-value=0.000

 reject H0
\item Model 'Jigsaw' VS 'Macro + Unlabeled'

 statistic=255.000, p-value=0.000

 reject H0
\item Model 'Jigsaw' VS 'Unlabeled'

 statistic=563.000, p-value=0.000

 reject H0
\item Model 'Jigsaw' VS 'Jigsaw + Unlabeled'

 statistic=1116.000, p-value=0.000

 reject H0
\item Model 'Jigsaw' VS 'Rot + Cutmix'

 statistic=1415.000, p-value=0.000

 reject H0
\item Model 'Jigsaw' VS 'Macro + Cutmix'

 statistic=580.000, p-value=0.000

 reject H0
\item Model 'Jigsaw' VS 'Jigsaw + Cutmix'

 statistic=1278.000, p-value=0.444

 Same proportions of errors (fail to reject H0)
\item Model 'Depth' VS 'Rotation'

 statistic=464.000, p-value=0.000

 reject H0
\item Model 'Depth' VS 'Macro'

 statistic=372.000, p-value=0.000

 reject H0
\item Model 'Depth' VS 'Jigsaw'

 statistic=1066.000, p-value=0.000

 reject H0
\item Model 'Depth' VS 'Hog'

 statistic=390.000, p-value=0.000

 reject H0
\item Model 'Depth' VS 'Rot + Unlabeled'

 statistic=265.000, p-value=0.000

 reject H0
\item Model 'Depth' VS 'Macro + Unlabeled'

 statistic=192.000, p-value=0.000

 reject H0
\item Model 'Depth' VS 'Unlabeled'

 statistic=453.000, p-value=0.000

 reject H0
\item Model 'Depth' VS 'Jigsaw + Unlabeled'

 statistic=1218.000, p-value=0.000

 reject H0
\item Model 'Depth' VS 'Rot + Cutmix'

 statistic=816.000, p-value=0.000

 reject H0
\item Model 'Depth' VS 'Macro + Cutmix'

 statistic=352.000, p-value=0.000

 reject H0
\item Model 'Depth' VS 'Jigsaw + Cutmix'

 statistic=857.000, p-value=0.000

 reject H0
\item Model 'Hog' VS 'Rotation'

 statistic=770.000, p-value=0.000

 reject H0
\item Model 'Hog' VS 'Macro'

 statistic=460.000, p-value=0.095

 Same proportions of errors (fail to reject H0)
\item Model 'Hog' VS 'Jigsaw'

 statistic=523.000, p-value=0.000

 reject H0
\item Model 'Hog' VS 'Depth'

 statistic=390.000, p-value=0.000

 reject H0
\item Model 'Hog' VS 'Rot + Unlabeled'

 statistic=560.000, p-value=0.000

 reject H0
\item Model 'Hog' VS 'Macro + Unlabeled'

 statistic=312.000, p-value=0.000

 reject H0
\item Model 'Hog' VS 'Unlabeled'

 statistic=774.000, p-value=0.800

 Same proportions of errors (fail to reject H0)
\item Model 'Hog' VS 'Jigsaw + Unlabeled'

 statistic=484.000, p-value=0.000

 reject H0
\item Model 'Hog' VS 'Rot + Cutmix'

 statistic=629.000, p-value=0.000

 reject H0
\item Model 'Hog' VS 'Macro + Cutmix'

 statistic=631.000, p-value=0.000

 reject H0
\item Model 'Hog' VS 'Jigsaw + Cutmix'

 statistic=445.000, p-value=0.000

 reject H0
\item Model 'Rot + Unlabeled' VS 'Rotation'

 statistic=395.000, p-value=0.000

 reject H0
\item Model 'Rot + Unlabeled' VS 'Macro'

 statistic=509.000, p-value=0.000

 reject H0
\item Model 'Rot + Unlabeled' VS 'Jigsaw'

 statistic=344.000, p-value=0.000

 reject H0
\item Model 'Rot + Unlabeled' VS 'Depth'

 statistic=265.000, p-value=0.000

 reject H0
\item Model 'Rot + Unlabeled' VS 'Hog'

 statistic=560.000, p-value=0.000

 reject H0
\item Model 'Rot + Unlabeled' VS 'Macro + Unlabeled'

 statistic=470.000, p-value=0.000

 reject H0
\item Model 'Rot + Unlabeled' VS 'Unlabeled'

 statistic=475.000, p-value=0.000

 reject H0
\item Model 'Rot + Unlabeled' VS 'Jigsaw + Unlabeled'

 statistic=310.000, p-value=0.000

 reject H0
\item Model 'Rot + Unlabeled' VS 'Rot + Cutmix'

 statistic=261.000, p-value=0.000

 reject H0
\item Model 'Rot + Unlabeled' VS 'Macro + Cutmix'

 statistic=460.000, p-value=0.000

 reject H0
\item Model 'Rot + Unlabeled' VS 'Jigsaw + Cutmix'

 statistic=326.000, p-value=0.000

 reject H0
\item Model 'Macro + Unlabeled' VS 'Rotation'

 statistic=380.000, p-value=0.000

 reject H0
\item Model 'Macro + Unlabeled' VS 'Macro'

 statistic=287.000, p-value=0.000

 reject H0
\item Model 'Macro + Unlabeled' VS 'Jigsaw'

 statistic=255.000, p-value=0.000

 reject H0
\item Model 'Macro + Unlabeled' VS 'Depth'

 statistic=192.000, p-value=0.000

 reject H0
\item Model 'Macro + Unlabeled' VS 'Hog'

 statistic=312.000, p-value=0.000

 reject H0
\item Model 'Macro + Unlabeled' VS 'Rot + Unlabeled'

 statistic=470.000, p-value=0.000

 reject H0
\item Model 'Macro + Unlabeled' VS 'Unlabeled'

 statistic=225.000, p-value=0.000

 reject H0
\item Model 'Macro + Unlabeled' VS 'Jigsaw + Unlabeled'

 statistic=191.000, p-value=0.000

 reject H0
\item Model 'Macro + Unlabeled' VS 'Rot + Cutmix'

 statistic=292.000, p-value=0.000

 reject H0
\item Model 'Macro + Unlabeled' VS 'Macro + Cutmix'

 statistic=213.000, p-value=0.000

 reject H0
\item Model 'Macro + Unlabeled' VS 'Jigsaw + Cutmix'

 statistic=201.000, p-value=0.000

 reject H0
\item Model 'Unlabeled' VS 'Rotation'

 statistic=788.000, p-value=0.000

 reject H0
\item Model 'Unlabeled' VS 'Macro'

 statistic=715.000, p-value=0.103

 Same proportions of errors (fail to reject H0)
\item Model 'Unlabeled' VS 'Jigsaw'

 statistic=563.000, p-value=0.000

 reject H0
\item Model 'Unlabeled' VS 'Depth'

 statistic=453.000, p-value=0.000

 reject H0
\item Model 'Unlabeled' VS 'Hog'

 statistic=774.000, p-value=0.800

 Same proportions of errors (fail to reject H0)
\item Model 'Unlabeled' VS 'Rot + Unlabeled'

 statistic=475.000, p-value=0.000

 reject H0
\item Model 'Unlabeled' VS 'Macro + Unlabeled'

 statistic=225.000, p-value=0.000

 reject H0
\item Model 'Unlabeled' VS 'Jigsaw + Unlabeled'

 statistic=452.000, p-value=0.000

 reject H0
\item Model 'Unlabeled' VS 'Rot + Cutmix'

 statistic=574.000, p-value=0.000

 reject H0
\item Model 'Unlabeled' VS 'Macro + Cutmix'

 statistic=622.000, p-value=0.000

 reject H0
\item Model 'Unlabeled' VS 'Jigsaw + Cutmix'

 statistic=482.000, p-value=0.000

 reject H0
\item Model 'Jigsaw + Unlabeled' VS 'Rotation'

 statistic=638.000, p-value=0.000

 reject H0
\item Model 'Jigsaw + Unlabeled' VS 'Macro'

 statistic=472.000, p-value=0.000

 reject H0
\item Model 'Jigsaw + Unlabeled' VS 'Jigsaw'

 statistic=1116.000, p-value=0.000

 reject H0
\item Model 'Jigsaw + Unlabeled' VS 'Depth'

 statistic=1218.000, p-value=0.000

 reject H0
\item Model 'Jigsaw + Unlabeled' VS 'Hog'

 statistic=484.000, p-value=0.000

 reject H0
\item Model 'Jigsaw + Unlabeled' VS 'Rot + Unlabeled'

 statistic=310.000, p-value=0.000

 reject H0
\item Model 'Jigsaw + Unlabeled' VS 'Macro + Unlabeled'

 statistic=191.000, p-value=0.000

 reject H0
\item Model 'Jigsaw + Unlabeled' VS 'Unlabeled'

 statistic=452.000, p-value=0.000

 reject H0
\item Model 'Jigsaw + Unlabeled' VS 'Rot + Cutmix'

 statistic=1305.000, p-value=0.000

 reject H0
\item Model 'Jigsaw + Unlabeled' VS 'Macro + Cutmix'

 statistic=463.000, p-value=0.000

 reject H0
\item Model 'Jigsaw + Unlabeled' VS 'Jigsaw + Cutmix'

 statistic=968.000, p-value=0.000

 reject H0
\item Model 'Rot + Cutmix' VS 'Rotation'

 statistic=610.000, p-value=0.000

 reject H0
\item Model 'Rot + Cutmix' VS 'Macro'

 statistic=621.000, p-value=0.000

 reject H0
\item Model 'Rot + Cutmix' VS 'Jigsaw'

 statistic=1415.000, p-value=0.000

 reject H0
\item Model 'Rot + Cutmix' VS 'Depth'

 statistic=816.000, p-value=0.000

 reject H0
\item Model 'Rot + Cutmix' VS 'Hog'

 statistic=629.000, p-value=0.000

 reject H0
\item Model 'Rot + Cutmix' VS 'Rot + Unlabeled'

 statistic=261.000, p-value=0.000

 reject H0
\item Model 'Rot + Cutmix' VS 'Macro + Unlabeled'

 statistic=292.000, p-value=0.000

 reject H0
\item Model 'Rot + Cutmix' VS 'Unlabeled'

 statistic=574.000, p-value=0.000

 reject H0
\item Model 'Rot + Cutmix' VS 'Jigsaw + Unlabeled'

 statistic=1305.000, p-value=0.000

 reject H0
\item Model 'Rot + Cutmix' VS 'Macro + Cutmix'

 statistic=561.000, p-value=0.000

 reject H0
\item Model 'Rot + Cutmix' VS 'Jigsaw + Cutmix'

 statistic=1275.000, p-value=0.000

 reject H0
\item Model 'Macro + Cutmix' VS 'Rotation'

 statistic=842.000, p-value=0.010

 reject H0
\item Model 'Macro + Cutmix' VS 'Macro'

 statistic=631.000, p-value=0.010

 reject H0
\item Model 'Macro + Cutmix' VS 'Jigsaw'

 statistic=580.000, p-value=0.000

 reject H0
\item Model 'Macro + Cutmix' VS 'Depth'

 statistic=352.000, p-value=0.000

 reject H0
\item Model 'Macro + Cutmix' VS 'Hog'

 statistic=631.000, p-value=0.000

 reject H0
\item Model 'Macro + Cutmix' VS 'Rot + Unlabeled'

 statistic=460.000, p-value=0.000

 reject H0
\item Model 'Macro + Cutmix' VS 'Macro + Unlabeled'

 statistic=213.000, p-value=0.000

 reject H0
\item Model 'Macro + Cutmix' VS 'Unlabeled'

 statistic=622.000, p-value=0.000

 reject H0
\item Model 'Macro + Cutmix' VS 'Jigsaw + Unlabeled'

 statistic=463.000, p-value=0.000

 reject H0
\item Model 'Macro + Cutmix' VS 'Rot + Cutmix'

 statistic=561.000, p-value=0.000

 reject H0
\item Model 'Macro + Cutmix' VS 'Jigsaw + Cutmix'

 statistic=327.000, p-value=0.000

 reject H0
\item Model 'Jigsaw + Cutmix' VS 'Rotation'

 statistic=590.000, p-value=0.000

 reject H0
\item Model 'Jigsaw + Cutmix' VS 'Macro'

 statistic=468.000, p-value=0.000

 reject H0
\item Model 'Jigsaw + Cutmix' VS 'Jigsaw'

 statistic=1278.000, p-value=0.444

 Same proportions of errors (fail to reject H0)
\item Model 'Jigsaw + Cutmix' VS 'Depth'

 statistic=857.000, p-value=0.000

 reject H0
\item Model 'Jigsaw + Cutmix' VS 'Hog'

 statistic=445.000, p-value=0.000

 reject H0
\item Model 'Jigsaw + Cutmix' VS 'Rot + Unlabeled'

 statistic=326.000, p-value=0.000

 reject H0
\item Model 'Jigsaw + Cutmix' VS 'Macro + Unlabeled'

 statistic=201.000, p-value=0.000

 reject H0
\item Model 'Jigsaw + Cutmix' VS 'Unlabeled'

 statistic=482.000, p-value=0.000

 reject H0
\item Model 'Jigsaw + Cutmix' VS 'Jigsaw + Unlabeled'

 statistic=968.000, p-value=0.000

 reject H0
\item Model 'Jigsaw + Cutmix' VS 'Rot + Cutmix'

 statistic=1275.000, p-value=0.000

 reject H0
\item Model 'Jigsaw + Cutmix' VS 'Macro + Cutmix'

 statistic=327.000, p-value=0.000

 reject H0

\end{itemize}

\end{multicols}

The test is particularly suggested when the methods being compared can only be assessed once, e.g. on a single test set, as opposed to numerous evaluations using a resampling methodology, such as k-fold cross-validation. 

\subsection{GAT on a supplementary Chest X-ray dataset: NIH}
\label{appendixC2}
\begin{figure*}

\begin{center}
\centerline{\includegraphics[width=0.5\textwidth]{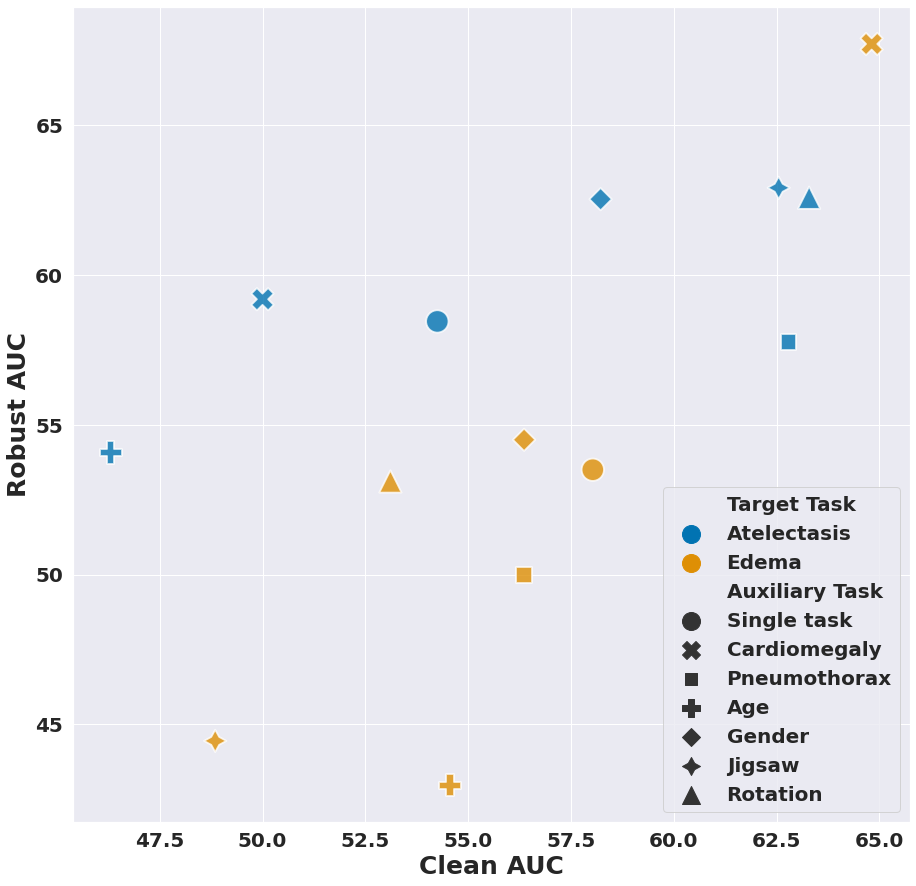}}

\caption{Comparison of different Task Augmentation strategies with single-task models using
Adversarial Training; Clean and robust AUC of GAT vs Single task adversarial training to diagnose Atelectasis and Edema pathologies for the NIH dataset}

\label{fig:nih}
\end{center}

\end{figure*}

We present in Figure \ref{fig:nih} similar study of the main paper, but on the NIH dataset.
Our conclusions that GAT outperforms Adversarial training (circles in \ref{fig:nih}) are confirmed on this dataset as well.

\begin{table*}[]
\begin{center}
\caption{Four Different $\mathcal{T}_i$ Scenarios: $\mathcal{T}_1$ ; standard training, $\mathcal{T}_2$ : adversarial training @ Goodfellow, $\mathcal{T}_3$ : adversarial training @ Madry, $\mathcal{T}_4$ : adversarial training @ Trades \cite{zhang2019trades}, and $\mathcal{T}_5$ : adversarial training @ Fast\cite{wong2020fast}, with 3 different task augmentations and equal weighting strategies.
} 
\label{table:full_adv_weight_strategies_appendix}
\begin{threeparttable}
\begin{tabular}{l|r|r|r|r|r|r}
\hline
Scenario  & \multicolumn{3}{c|}{Clean accuracy (\%)} & \multicolumn{3}{c}{Robust accuracy (\%)} \\

 &  \textit{Jigsaw} &   \textit{Macro} & \textit{Rotation} &   \textit{Jigsaw} &   \textit{Macro} & \textit{Rotation} \\
\hline
$\mathcal{T}_1$ : Standard  training   &  88.78 &  93.04 &    69.67 &   0.59 &  0.06 &     3.18 \\
\hline

$\mathcal{T}_3$ : Madry AT &  64.9 &  \textbf{83.00} &  68.23   &  20.25  &  32.16 &    25.43 \\

$\mathcal{T}_2$ : Goodfellow AT    &  55.01 &  77.49 &  42.29   &  40.5  &  38.24 &    34.43 \\

$\mathcal{T}_4$ : Trades  AT &  46.4 &  60.73 &  50.24    &  33.61  &  \textbf{42.76} &    42.05 \\

$\mathcal{T}_5$ : Fast AT &  52.35 &  78.18 &  56.36    &  19.06  &  26.56 &    19.84 \\
\bottomrule
\end{tabular}

\end{threeparttable}
\end{center}
\vspace{-3mm}
\end{table*}

\subsection{GAT combined with other weighting strategies}
\label{appendixWeight}
We evaluate 5 weighting strategies on Resnet-50 architectures:
\begin{enumerate}
    \item  Equal weights (Equal),
    \item  Impartial Multi-task Learning (IMTL) \cite{weight_imtl},
    \item  Multiple Gradient Descent Algorithm (MGDA) \cite{weight_mgda}, 
    \item  Gradient Vaccine (GradVac) \cite{weight_gv}, 
    \item  Project Conflicting Gradients (PCGrad) \cite{weight_pcg}
\end{enumerate}

The results in Table \ref{table:full_adv_weight_supp} uncover that adversarial training using the Macro task yields the best performance in 4 over 5 weighting strategies, and that MGDA weighting strategies yields the best clean and robust accuracy among the weighting strategies (with the \textbf{MACRO} task). MGDA uses multi-objective optimization to converge to the pareto-stationnary for both the tasks we train over. This search algorithm shows that we can attain loss landscapes with high clean and robust performances that greedy gradient algorithms (equal weights, GradVac, PCGrad) fail to uncover.     

We used the default hyper-parameters for the weighting strategies. One possible work would be to fine-tune the weighting strategies to the adversarial training setting.

\begin{table*}[]
\begin{center}
\caption{Evaluation results of Two Different $\mathcal{T}_i$ Scenarios: $\mathcal{T}_1$ (standard training), $\mathcal{T}_2$ (adversarial training), with 3 different task augmentations and 5 weighting strategies. In bold, the best values for each scenario } 
\label{table:full_adv_weight_supp}
\begin{threeparttable}
\begin{tabular}{c|l|r|r|r|r|r|r}
\hline
\multirow{2}{*}{Scenario} & \multirow{2}{*}{Weight} & \multicolumn{3}{c|}{Clean accuracy (\%)} & \multicolumn{3}{c}{Robust accuracy (\%)} \\

&  &  \textit{Jigsaw} &   \textit{Macro} & \textit{Rotation} &   \textit{Jigsaw} &   \textit{Macro} & \textit{Rotation} \\
\hline
\multirow{5}{*}{$\mathcal{T}_1$}& Equal     &  88.78 &  93.04 &    69.67 &   0.59 &  0.06 &     3.18 \\
& GradVac   &  89.08 &  93.01 &    68.42 &   0.26 &  0.09 &     3.81 \\
& IMTL      &  61.46 &  93.75 &    71.24 &   0.98 &  0.09 &     3.81 \\
& GAT [Ours]      &  41.65 &  \textbf{93.89} &    70.26 &   0.00 &  0.24 &     \textbf{4.33} \\
& PCGrad    &  88.85 &  92.99 &    69.11 &   0.69 &  0.11 &     3.13 \\
\hline
\multirow{5}{*}{$\mathcal{T}_2$}& Equal     &  55.01 &  \textbf{77.49} &  42.29   &  40.5  &  38.24 &    34.43 \\
& GradVac   &  44.67 &  64.11 &   57.71 &   36.24 &  35.56 &   40.17 \\
& IMTL      &  42.05 &  69.63 &   59.61 &   33.84 &  48.21 &    39.93 \\
& MGDA [Ours]     &  43.99 &  73.7 &    56.51 &   32.95 &  \textbf{48.38} &     36.13 \\
& PCGrad    &  41.6 &  63.76 &   56.38 &   33.8 & 44.74 & 41.59 \\

\bottomrule
\end{tabular}

\end{threeparttable}
\end{center}
\end{table*}

\begin{figure}
    \centering
    \includegraphics[width=0.5\textwidth]{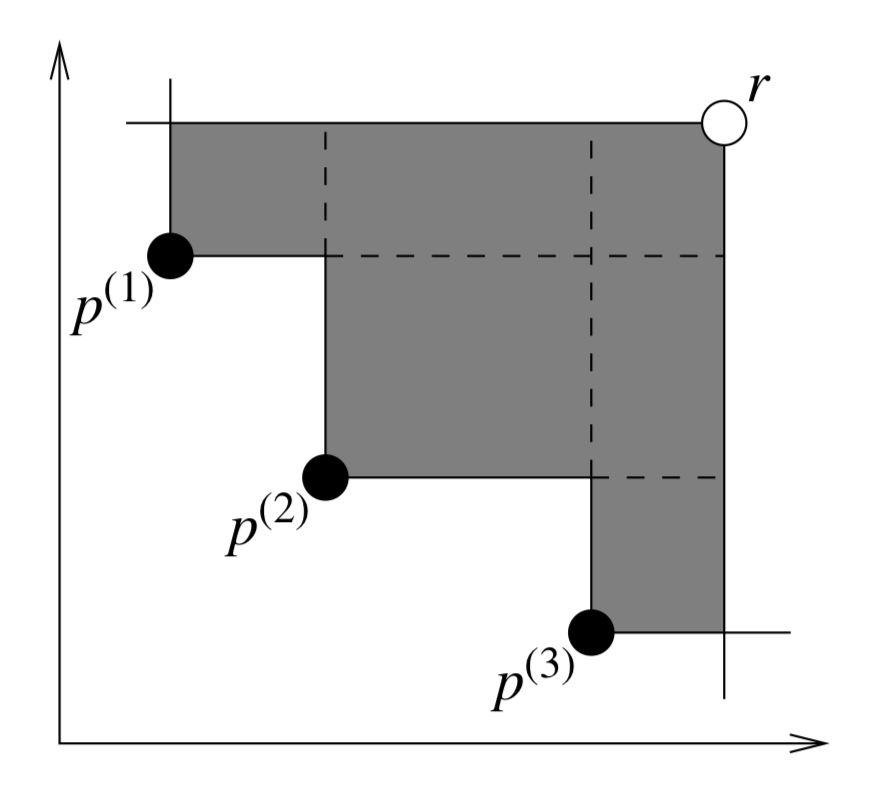}
    \caption{Hypervolume \cite{hypervolume}}
    \label{fig:hypervolume}
\end{figure}

\paragraph{Hyper-volume.} Hyper-volume is a popular metric to compare different pareto fronts. It only needs a reference point as showed in Fi.\ref{fig:hypervolume}. It calculates the area/volume, which is dominated by the provided set of solutions with respect to a reference point. We use the implementation from the Pymoo library\footnote{https://pymoo.org/misc/indicators.html}.

We provide in Table \ref{table:hypervolume} the hyper-volume metric of the fronts obtained using each of the weighting strategies (MGDA, IMTL, GV, PCG) we compare. Lower values indicate better solutions. The results confirm that using MGDA for GAT leads to better pareto-fronts.

\begin{table}[]
\centering
\caption{Hyper-volume of different fronts. Lower values mean better pareto-fronts. } 
\label{tab:full_adv_weight_hypervolume}
\begin{threeparttable}
\begin{tabular}{c|c}
\hline
Weight & Hyper-volume \\
\hline
MGDA (OURS)	& 0.4442\\
IMTL 	&0.4534\\
PCG 	&0.4679\\
GV 	    &0.4497 \\
 \bottomrule
\end{tabular}

\end{threeparttable}
\end{table}

\subsection{Impact of number of tasks}

We show in Table \ref{tab:full_adv_nb_tasks} how robust accuracy of models fluctuate depending on the choice and number of auxiliary tasks.
It seems that the Jigsaw task is the most vulnerable and causes significant degradation of the robustness of the model.

\begin{table}[]
\centering
\caption{Impact of the number of tasks on the robust accuracy of models.} 
\begin{tabular}{l|r|r}
\hline
Added tasks                        & Robust accuracy (\%) & Number of tasks \\
\hline
Target (No auxiliary task)         & 39.09               & 1               \\
Target + Jigsaw                    & 32.95               & 2               \\
Target + Rotation                  & 36.13               & 2               \\
Target + Macro                     & 48.38               & 2               \\
Target + Jigsaw + Rotation         & 35.45               & 3               \\
Target + Jigsaw + Macro            & 16.13               & 3               \\
Target + Rotation + Macro          & 56.21               & 3               \\
Target + Macro + Rotation + Jigsaw & 28.24               & 4               \\ \hline
\end{tabular}
\label{tab:full_adv_nb_tasks}
\end{table}

\subsection{Adaptive attacks: AutoAttack}

We evaluate for all the models of the study the adversarial robustness against AutoAttack. For 3/4 scenarios, adversarial training with task augmentation using \textbf{Macro} tasks outperforms single-task adversarial training.

\begin{table*}[h]
\begin{center}
\caption{Robust accuracy (\%) against AutoAttack of different models adversarially trained with GAT, with 3 different task augmentations, compared to their counterpart single task adversarially trained models. In bold the cases where GAT outperforms single-task AT.
} 
\label{table:generalization_supp}
\begin{threeparttable}
\begin{tabular}{r||lrrrr}
\hline
\multirow{2}{*}{Dataset} & \multirow{2}{*}{Scenario} & \multicolumn{4}{c}{Auxiliary task} \\
& &  \textit{None} &  \textit{Jigsaw} &   \textit{Macro} & \textit{Rotation} \\
\hline
\multirow{3}{*}{CIFAR-10} & 100\% Dataset & 27.01 & \textbf{29.63}  &  \textbf{32.54} &    13.82 \\
\cline{2-6}
 &  10\% Dataset & 14.27 & 11.63 & \textbf{15.00} & 11.56 \\
\cline{2-6}
 & WideResnet28-10  &  36.29 & 15.99 & 34.44 & 25.72 \\
 \cline{2-6}
 & WideResNet-70-16  &  36.29 & 15.99 & 34.44 & 25.72 \\
\hline
STL-10 & 100\% Dataset & 19.40 & 12.78 & \textbf{20.02} & 17.64 \\

\hline
\bottomrule
\end{tabular}

\end{threeparttable}
\end{center}
\vspace{-3mm}
\end{table*}

\subsection{Surrogate attacks}

We evaluate in Table \ref{table:full_adv_transfer} the transferability of attacks from a surrogate model to a target model. Both models are trained on the same training dataset. 

(1) When the target model has an auxiliary task, the success rate of the attack crafted from a single-task surrogate model drops by 14\%. (2) When the surrogate model has an auxiliary task, the success rate against a single task target model drops by 60\%. 

(1) indicates that the adversarial examples generated to fool a multi-task model actually lie in a loss landscape that is not adversarial for the single task model: The PGD optimization is misguided when multiple tasks are present.

(2) The adversarial examples generated against one single task are actual relevant to models with multiple tasks. It means that multitask learning by itself has the same vulnerable area as the single task-learning, it is just that gradient-based attacks have more difficulty to find them.

\begin{table*}[t]
\begin{center}
\caption{Evaluation results of Three Different combinations of surrogate models and target models. For each combination, we craft the adversarial examples on the surrogate and evaluate the success rate of the examples on the target models. Both surrogate and target models are trained with standard training.} 
\label{table:full_adv_transfer}
\begin{threeparttable}
\begin{tabular}{c|c|c|c}
\hline
Target $\rightarrow$ & Single Task & Auxiliary \textit{Rotation} & Auxiliary \textit{Jigsaw}  \\
\hline
Surrogate $\downarrow$ &  \multicolumn{3}{c}{Success rate \textit{\%}}  \\ \hline
Single Task & 98.47 & 84.55 & 86.56\\
Auxiliary \textit{Rotation} & 37.95 & 98.05 &  79.86\\
Auxiliary \textit{Jigsaw}  & 37.48 & 79.35 &  98.59\\
\hline
\end{tabular}
\end{threeparttable}
\end{center}
\vspace{-3mm}
\end{table*}


\end{document}